\documentclass{article}
\usepackage[utf8]{inputenc} 
\usepackage[T1]{fontenc}    
\usepackage{hyperref}       
\usepackage{url}            
\usepackage{booktabs}       
\usepackage{amsfonts}       
\usepackage{nicefrac}       
\usepackage{microtype}      
\usepackage{xcolor}         

\usepackage{enumitem}   
\usepackage{graphicx,xcolor}
\usepackage{epsfig}
\usepackage{amsmath}
\usepackage{amssymb}
\usepackage{caption}
\usepackage{mathtools}
\mathtoolsset{showonlyrefs}

\newcommand{\commentout}[1]{}

\def \d {{\mathrm d}}

\def \Rset {{\mathbb R}}

\def \Nset {{\mathbb N}}

\newcommand{\be}{\begin{equation}}
\newcommand{\ee}{\end{equation}}
\newcommand{\ba}{\begin{eqnarray}}
\newcommand{\ea}{\end{eqnarray}}
\newcommand{\bi}{\begin{itemize}}
\newcommand{\ei}{\end{itemize}}
\newcommand{\br}{\begin{eqnarray}}
\newcommand{\er}{\end{eqnarray}}

\newcommand{\qed}{\mbox{$\square$}\newline}

\newtheorem{theo}{Theorem}[section]

\newtheorem{example}[theo]{Example}

\newtheorem{prop}[theo]{Proposition}
\newtheorem{lem}[theo]{Lemma}
\newtheorem{cor}[theo]{Corollary}
\newtheorem{rmk}[theo]{Remark}
\newtheorem{assp}[theo]{Assumption}

\begin{document}

\title{{\Large\bf{Global Well-posedness and Convergence Analysis of Score-based Generative Models via Sharp Lipschitz Estimates}}}

\author{Connor Mooney\thanks{mooneycr@math.uci.edu. CM was partly supported by Alfred P. Sloan Research Fellowship, NSF CAREER grant DMS-2143668.}, Zhongjian Wang\thanks{(Corresponding) zhongjian.wang@ntu.edu.sg.   ZW was supported in part by NTU SUG-023162-00001.},  Jack Xin\thanks{jack.xin@uci.edu. JX was partly supported by NSF grants DMS-2151235, DMS-2219904, DMS-2309520.},  Yifeng Yu\thanks{yifengy@uci.edu. YY was partly supported by NSF grant DMS-2000191.}}

\date{}
\maketitle

\begin{abstract}
We establish global well-posedness and convergence of the score-based generative models (SGM) under minimal general assumptions of initial data for score estimation. \textit{For the smooth case}, we start from a Lipschitz bound of the score function with optimal time length. The optimality is validated by an example whose Lipschitz constant of scores is bounded at initial but blows up in finite time. This necessitates the separation of time scales in conventional bounds for non-log-concave distributions. In contrast, our follow up analysis only relies on a local Lipschitz condition and is valid globally in time. This leads to the convergence of numerical scheme without time separation.  \textit{For the non-smooth case,} we show that the optimal Lipschitz bound is $O(1/t)$ in the point-wise sense for distributions supported on a compact, smooth and low-dimensional manifold with boundary. 
\end{abstract}

\section{Introduction}
Diffusion models (DM) have become the state-of-the-art tools lately in generative AI 
\cite{Gen_grad_datadistr_2019,ScoreDM_2021,DMbGAN2021} such as image synthesis \cite{CasDM_2022,MDT2023}.
DMs first evolve data samples with stochastic differential equation (SDE) to gradually inject Gaussian noise until a Gaussian distribution is reached. Then it approximates the drift in the associated backward (time-reversed) SDE and 
generate a data sample from Gaussian noise.
The drift of the backward SDE contains
the gradient of the forward logarithmic density (score) that is estimated by solving a matching problem with deep neural network training. The reversibility concept of SDEs
dated back to Kolmogorov's work \cite{K37} in 1937, and the general score formula was derived by Anderson \cite{And_80} in 1982. 

Theoretical study on the convergence of DM generated distribution to the target (data) distribution typically assumes that the data distribution admits a density with respect to Lebesgue measure \cite{lee2022convergence} among others. By also imposing that the {\it score
of the data distribution is Lipschitz continuous}, the score function of the forward process (the drift in the backward process) is well-behaved (not exploding) as the backward time tends to zero when the desired target sample is to be generated. However, this is not always observed in
practice and experimentally the score can blow up \cite{score_exploding_2022}. In particular, the explosion occurs at generation if the data distribution satisfies the {\it manifold hypothesis (MH)} \cite{ManiHyp_2020,Goodfellow_2016} which is verified for image data in \cite{UMHyp_DL_2023}.  Under MH, \cite{Pid_2022} 
showed that the limit of the continuous backward process with approximate score is well-defined and that the sample distribution shares the same support as the target distribution under the integrability conditions on the error of score matching.
Also under MH, \cite{B2023} found quantitative bounds on the 1-Wasserstein distance between a   compact target (data) distribution and the generative
distribution of DM by allowing the score function to explode as backward time approaches zero. 

Both of the references (\cite{Pid_2022}
 and \cite{B2023}), among others (\cite{chen2022sampling} for smooth case in score-based model, \cite{lee2022convergence} for Langevin MC, \cite{huang2024convergence} for ODE flows, \cite{chang2024deep} for F\"{o}llmer flows), require a (locally) Lipschitz estimate for the score function to ensure the well-poseness of the backward SDEs and the approximation bound of the score matching and sampling process. 

 \textbf{The goal of this paper} is to provide sharp estimates  that 1)  confirm/improve the score assumptions of the existing convergence theory, 2) give insight for the duration of the forward process so that the backward process is well-defined, and 3) justify practical implementation of the backward process (e.g. early stopping strategies or truncation \cite{score_exploding_2022}).

\paragraph{Related work} We are aware of the convergence bound of discrete schemes for backward processes in \cite{chen2023improved}. Our convergence bound 
takes the KL chain inequality (Proposition C.3.\cite{chen2023improved}) as the building block. While equipped with sharp (local) Lipschitz bounds in the paper, we achieved polynomial complexity of sampling in the general smooth $p_0$ setting without separated regimes of schedule. We are also aware of \cite{B2023} which provides convergence bound in Wasserstein distance under a singular $p_0$ setting, supported on a compact manifold. Due to the potential singular behaviour of the score, early stopping schedules are employed \cite{score_exploding_2022}. Our paper provides sharp Lipschitz bounds of the singularities and therefore insights for the choices of schedules and loss normalization between discretization points.

 \textbf{The main contributions of this paper are:}
 \begin{itemize}
     
 \item 
 {\it Realistic or sharp point-wise gradient and Hessian estimates} of 
the score potential function $\log p$ from commonly hypothesized data distributions. 

\item The first sharp example demonstrating the loss of Lipschtiz bound of the score function as time gets large even with nice initial data.

\item Well-posedness and convergence of the backward diffusion process up to time zero (the generation time) in the  smooth setting without separated regime of discretization.

\item Characterization of the score (and its derivatives) in the setting of manifold hypothesis.

\end{itemize}

The rest of the paper is organized as follows. In Section \ref{sec:prelim}, we first introduce settings of the diffusion model and discretization schemes of the backward process. Later, we present the transformation that relates the Fokker Planc equation with unbounded coefficients (density equation of forward process) to the non-linear Hamilton Jacobi equation and heat equation, which serves as the foundation of the analysis. The main theoretical results, Hessian estimate of score potential function $\log p$, are listed in Section \ref{sec:HessionEstimate}. Based on these estimates, we establish well-posedness of the continous backward process and convergence bound of discretization in Section \ref{sec:convergence}. The details of the proofs are in the Appendix.

\section{Preliminaries}\label{sec:prelim}
\subsection{Background and  Setting the Stage}
A large class of generative diffusion models can be analyzed under the SDE framework \cite{ScoreDM_2021}. It consists of two processes: forward and backward. \emph{The forward process}, which relates to training, is an Ornstein-Uhlenbeck (OU) process in $\Rset^n$ as follows:
\begin{equation}\label{forward}
    d X_t = -\frac{1}{2}X_t dt + dW_t, \quad \text{for $t\in [0,T]$}
\end{equation}
where $W_t$ is a standard Brownian motion, $T$ is the final time such that the distribution of $X_T$ approximates a normal distribution in $\Rset^n$, namely $\mathcal{N}(0,I_n)$. The initial distribution $X_0$ follows a target (data) distribution in $\Rset^n$ during the generative task, denoted as $p_0$.  \emph{The backward process}, which relates to generation of new data, is defined as an 'inversion' of forward process \eqref{forward}. More precisely, with time reversal $t'=T-t$,
\be\label{mainode}
d\tilde X_{t'}=\left({\frac{1}{2}}\tilde X_{t'}+\nabla\log p(T-t',\tilde X_{t'})\right)\,d{t'}+d\tilde W_{t'}\quad \text{for $t'\in [0,T]$},
\ee
where $W_{t'}$ is a standard Brownian motion (not necessarily being the same as $W_t$) and the initial distribution $\tilde X_0$ follows $\mathcal{N}(0,I_n)$. The term $\nabla\log p$ is introduced in Eq. \eqref{mainode} such that the marginal distributions of the forward and backward processes are identical \cite{And_80}. 

To be specific, let $p:=p(x,t)$ denote the probability distribution function of the forward process \eqref{forward}, which solves the Fokker Planck equation with Cauchy data $p_0$, namely
\be\label{fkeqn}
\begin{cases}
p_t=\frac{1}{2}(\nabla \cdot (xp)+\Delta \, p)\\
p(x,0)=p_0(x).
\end{cases}.
\ee
We also denote $P_t$ ($Q_t'$ correspondingly) as the marginal distribution of $X_t$ in \eqref{forward} ($\tilde X_{t'}$ in \eqref{mainode}). Given initial distribution for \eqref{mainode} $Q_0\sim P_T$, then \cite{And_80}: $\forall t$, $Q_t = P_{T-t}$. Especially, $Q_T=P_0$ so data  $\sim p_0$ can be generated by solving \eqref{mainode}.

In practice, since no closed form expression of $p_0$ is known, the $p$ in \eqref{fkeqn} is not analytically available. 
Thus $\nabla\log p$ is approximated by a neural network $s:=s_\theta(t,x)$, where $\theta$ denotes latent variables of neural network and is omitted for simplicity of notation. The approximation is obtained by training the neural network with an $L_2$ score estimation loss, $\forall t\in[0,T]$,
\begin{align*}
    \mathbb{E}_{x\sim P_t}||s_\theta(t,x)-\nabla \log p(t,x)||^2.
\end{align*}
In the analysis, we assume an $\epsilon^2$ bounds for this estimation, see Assumption \ref{ass:score}.

Given the approximation of score $s_\theta$, we employ the exponential  scheme \cite{zhang2022fast} with initial distribution $\mathcal{N}(0,I_n)$. More precisely, let $ \delta = t_0 \leq t_1 \leq \cdots \leq t_N = T$ be the discretization points. $\delta = 0$ for the normal setting and $\delta>0$ for the early-stopping setting. Then with $t_k'=T-t_{N-k}$, the process in the discrete scheme is as follows:
\begin{align}\label{semi-discrete}
    d \hat{x}_{t'} = (\frac{1}{2}\hat{x}_{t'}+s_\theta(\hat{x}_{t_k'},T-t_k'))dt +d\hat w_{t'} \quad t'\in[t_k',t_{k+1}'], \quad k=0,\cdots,N-1,
\end{align}
which admits an explicit solution, with $\mu_k \sim \mathcal{N}(0,I_n)$,
\begin{align}
    d\hat{x}_{t'_{k+1}}=e^{\frac{1}{2}(t'_{k+1}-t'_k)} \hat{x}_{t'_{k}} + 2(e^{\frac{1}{2}(t'_{k+1}-t'_k)}-1)s_\theta(\hat{x}_{t_k'},T-t_k'))+\sqrt{e^{(t'_{k+1}-t'_k)}-1} \mu_k .
\end{align}

Due to the limited knowledge of $p_0$ as well as the regularity of $\nabla \log p$, we restrict ourselves to  uniform discretization points. Detailed selection is stated in the convergence theorems. 

We assume the following bound of score approximation at the discretization points,
\begin{assp}\label{ass:score} Let $t_k$ be the discretization point of the scheme \eqref{semi-discrete}, 
    \begin{align} \label{score-error}
\frac{1}{T}\sum_{k=1}^N (t_k-t_{k-1})E_{x\sim P_{t_k}} \|\nabla \log p(t_k,x) - s(t_k,x) \|^2 \leq \epsilon_0^2.
\end{align}
\end{assp}

\subsection{Foundational Ideas based on Non-linear Hamilton Jacobi Equation }
The foundation of our analysis is investigating the behaviour of $\log p$ as the solution of a non-linear Hamilton Jacobi equation (HJE).
We consider the score potential function\footnote{Here we only consider the transform when the distribution of forward process $P_t$ is absolutely continuous with respect to Lebesgue measure. The transform and our analysis are valid for any $t>0$ in the general case and up to $t=0$ when $p_0$ is smooth. }
\begin{align}
    q(t,x)=-\log\, p(t,x)-\frac{|x|^2}{2}
\end{align}
whose spatial gradient becomes the drift (score) in the backward (reverse time denoising and generation) process \eqref{mainode} of the diffusion model. 
The $q$ function 
satisfies the following PDE: 
\begin{equation}\label{q-eqn}
\begin{cases}
     q_t-\frac{1}{2}\Delta q +\frac{1}{2} (x\cdot \nabla q  +|\nabla q|^2)=0
    \\
    q(0,x) = g(x), 
    \end{cases}
\end{equation}
where $g(x)= -\log p_0(x) - |x|^2/2$, which is the non-Gaussian part of the likelihood function.

To simplify Eq.\eqref{q-eqn},  we make a two step change of variables in time.  First, let $\tilde{q}(t,x)= q(t,e^{t/2}x)$, then $\tilde{q}$ solves:
\begin{align}
     \tilde q_t =q_t+e^{t\over 2}x\cdot \nabla q(t,e^{t\over 2}x)= \frac{e^{-t}}{2}(\Delta \tilde q -|\nabla \tilde q|^2).
\end{align}

   Then we consider $\bar{q}(t,x)=\tilde{q}(-\log(1-t),x)$, then $\bar{q}$ solves:
   \begin{equation}
   \begin{cases}
       \bar q_t =\frac{1}{2}(\Delta \bar q -|\nabla \bar q|^2) \quad t\in[0,1) \\
       \bar q (0,x)=q_0
   \end{cases}.
        \label{vHJ} 
   \end{equation}
\begin{rmk}\label{rmk:transform}
   By a direct calculation
\begin{align}\label{transform}
    \bar q(t,x)=q\left(-\log(1-t),\  \frac{1}{\sqrt{1-t}} x\right) \text{ or equivalently, }q(t,x)=\bar q(1-e^{-t},e^{-t/2}x).
\end{align}
Furthermore,
\begin{align}
    \nabla q(t,x) = e^{-t/2}\nabla \bar q(1-e^{-t},e^{-t/2}x) \text{ and, } \nabla^2 q(t,x) = e^{-t}\nabla \bar q(1-e^{-t},e^{-t/2}x).
\end{align} 
\end{rmk}

Lastly, we also define $\bar p(t,x)=e^{-\bar q(t,x)}$, which satisfies 
\begin{equation}\label{heat}
    \begin{cases}
\bar p_t={\frac{1}{2}}\Delta \bar p \quad \text{on $(0,1)\times \Rset^n$}\\
\bar p(0,x)=h(x)=e^{-g(x)}.
\end{cases}.
\end{equation}

The solution of  \eqref{heat} is given by 
$$
\bar p(x,t)={1\over (2\pi t)^{n\over 2}}\int_{\Rset^n}e^{-|x-y|^2\over 2t}e^{-g(y)}\,dy.
$$

To derive reasonable point-wise estimates of gradients and Hessian of the score function  $q(t,x)$ that does not involve $1/ t$, we will need the following assumption in relevant results. This assumption also ensures the above integration is well defined for $t\in [0,1]$.
\begin{assp}\label{assp:max-logconvx} The tail distribution is bounded by some Gaussian distribution, i.e,
\begin{align}
\log p_0(x)-\log p_0(0) \leq \alpha_1+\frac{1}{2}(\alpha_2-1)|x|^2
\end{align}
for constants 
 $ \alpha_2<1$ and $\alpha_1 \in R $. Without loss of generality we assume $\alpha_2>0$. 
\end{assp}
Recalling definition of $g$, it is equivalent to 
\begin{align}
    \label{generalcondition}
g(x)-g(0)\geq -{\alpha_2\over 2} |x|^2-\alpha_1,
\end{align}
Technically speaking,  the $g(0)$ could be absorbed into $\alpha_1$. We put it there just to track possible dependence on the dimension $n$. 

Note that Assumption \ref{assp:max-logconvx} implies that the second order moment of the process is bounded, i.e.,  $E_{p_0}||X||^2=M_2< \infty$.

\paragraph{General notations}
Throughout this paper,  for an $n\times n$ matrix $A$, we use the spectral norm
\be\label{matrixnorm}
||A||_2=\max_{\{v\in \Rset^n:\ |v|=1\}}|Av|=\text{the largest eigenvalue of $\sqrt{AA^{\top}}$}.
\ee
In particular, for a  map $F:\Rset^n\to \Rset^n$,
$$
||\nabla F||_2\leq L \quad \Leftrightarrow \quad |F(x)-F(y)|\leq L|x-y|.
$$
We also adopt the following notation when comparing two symmetric (Hessian) matrices,
\begin{align}
A \preceq B  \text{ if $B-A$ is semi-positive definite.} 
\end{align}
So for any symmetric matrix $A$, $\|A\|_2\leq \sigma \Leftrightarrow -\sigma I_n\preceq A\preceq\sigma I_n$.

\section{Sharp Hessian Bound of Score Potential Function}
\label{sec:HessionEstimate}

The fundamental question, which is directly related to the well-posedness and convergence rate of the diffusion model \cite{B2023,lee2022convergence},  is whether for any $T>0$, there exists a constant $C_T$ that depends only on $T$ and the initial data such that
\be\label{target}
\sup_{[0,T]\times  \Rset^n}\|D_x^2q(t,x)\|_2\leq C_T\ ?
\ee
In Section \ref{sec:H-est} we provide a uniform Hessian bound which only lasts for a finite time if the $p_0$ is not log-concave. Interestingly,  the bound turns out to be optimal in sense of lasting time as in Example \ref{eg:counter-eg} we construct an initial distribution $p_0$ such that the Hessian of $\log p$ loses global bound {\bf right} at the limiting time.
Inspired by the counter-example, alternatively in Section \ref{sec:local-est} we provide a locally Lipschitz estimate that lasts for $t\in[0,\infty)$. For the non-smooth case, in Section \ref{sec:compact}, we characterize the singular behaviour of $\log p$ and its derivatives.

\subsection{Hessian Estimate of Score Potential Function in Finite Time}\label{sec:H-est}
\begin{theo}\label{secondestimate} Let $M_0$ be a nonnegative number. $g\in C^2(\Rset^d)$\footnote{The assumption is equivalent to $\log p_0\in C^2(\Rset^d)$.}. 

(1)  If $D^2g(x)\preceq M_1I_n$, then
$$
D_x^2q(t,x)\preceq e^{-{t}}M_1I_n \quad \text{for all $(t,x)\in [0,\infty)\times \Rset^n$}.
$$

(2) If $D^2g(x)\succeq -M_0I_n$, then for any $T\in \left[0, -\log(1-\frac{1}{M_0})\right)$, we have 
$$
D_x^2q(t,x)\succeq -\frac{M_0}{e^t-M_0(e^{t}-1)}I_n \quad \text{for all $(t,x)\in (0,T]\times \Rset^n$}.
$$
Note that if $M_0\leq 1$, then $T\in [0,\infty)$. 
\end{theo}

The proof is in Section \ref{proof:secondestimate}.As an immediate corollary,  we have that
\begin{cor}\label{Hessianbound} Given data distribution $p_0\in C^2(\Rset^d)$ follows $-L_1 I \preceq\sup_{x\in \Rset^n}D^2\log p_0(x)\preceq L_0 I$.  Then we have short time uniform bound of the Hessian:  for any $t\in \left[0, -\log(1-\frac{1}{L_0+1})\right)$,
$$
\sup_{ \Rset^n}\|D_x^2\log p(t,x)\|_2\leq C_t. 
$$
where $C_t=\max\left(\frac{L_0+1}{1-(L_0+1)(e^t-1)}-1,\ e^{-t}(L_1-1)+1\right)$. 

 Furthermore, if $-\log p_0(x)$ is a convex function ($L_0\leq 0$), the estimate  bound is global,
$$
0\preceq -D_x^2\log p(t,x)\preceq (e^{-{t}}L_1+(1-e^{-t}))I_n \quad \text{for all $(t,x)\in [0,\infty)\times \Rset^n$}.
$$
\end{cor}
\begin{rmk}
    The convex case has been also discussed in \cite{lee2021universal} and it leads to single modal distribution.   Similar finite bound was also derived in Lemma C.9 in \cite{chen2023improved}, which follows directly from the expression formula and the generalized Poincar\'e inequality for log-concave probability measures. Our Theorem \ref{2ndorderestimate} can be viewed as an improved version since the upper bound of $D^2\log p(t,x)$ only depends on the upper bound of  $D^2\log p_0(x)$ instead of $||D^2\log p_0(x)||_2$ as in  \cite{chen2023improved}. The PDE (partial differential equation) approach in our proof is more robust, which does not rely on the formula and could be easily adjusted to more general situations. 
\end{rmk}

In the following,  we show that the temporal bound $-\log\left(1-\frac{1}{M_0}\right)$ in the statements of  Theorem \ref{secondestimate} and Corollary \ref{Hessianbound} is sharp.\begin{example}[Loss of Uniform Hessian Bound]\label{eg:counter-eg}
There  exists a smooth nonnegative $g$ satisfying assumptions in Theorem \ref{secondestimate} and Corollary \ref{Hessianbound} ($M_0=M_1=2$) such that the corresponding $q(x,t)$ satisfies 
$$
\sup_{x\in \Rset^n}||D^2 q\left(\log 2,x\right)||=\sup_{x\in \Rset^n}||D^2\bar q\left({1/2},x\right)||=\infty. 
$$

\end{example}

Note that the number $1\over 2$ can be changed to any given time by re-scaling the function $\bar q(\lambda^2 t, \lambda x).$ The detail of construction is in Section \ref{proof:counter-eg}

\begin{rmk}We are aware of some existing literature where similar finite time Hessian estimate is obtained given regularity of the initial data, for instance Lemma C.9. in \cite{chen2023improved} and \cite{brigati2024heat}.  Our result, as stated in Theorem  \ref{secondestimate} is sharper in terms of longer valid time and consistency when $t\to 0$. More precisely, the key ingredient of the analysis in \cite{chen2023improved} and in Theorem  \ref{secondestimate} is the 'semi-convexity' of $-\nabla \log p$. For convex $p_0$, our analysis is valid for any compact interval of $[0,\infty)$ while \cite{chen2023improved} lasts for $t$ such that, $
    e^{\frac{t}{2}}-e^{-\frac{t}{2}}\leq \frac{1}{2L},
$ where $L$ is the Lipschitz constant of $\nabla \log p_0$. For non-convex $p_0$, taking example under construction, the Lipschitz constant $L=M_0+1$. Then the upper bound 
turns to $e^{\frac{t}{2}}-e^{-\frac{t}{2}}\leq \frac{1}{2M_0+2}$. In Theorem  \ref{secondestimate}, it turns to
\begin{align}\label{eqn:com-jf-est}
    e^{-t}\geq 1-\frac{1}{M_0}&\Leftrightarrow (e^{\frac{t}{2}}-e^{-\frac{t}{2}})^2\leq \frac{1}{M_0(M_0-1)}
    \Leftarrow e^{\frac{t}{2}}-e^{-\frac{t}{2}}\leq \frac{1}{2M_0+2},
\end{align}
 which indicates our bound is better than  \cite{chen2023improved}.
Furthermore, our estimate of upper bound of time does not require the upper bound of $D^2(-\log p)$.
\end{rmk}
\subsection{Local Estimate}\label{sec:local-est}

The following theorem provides point-wise estimates of the score function, which can be quite useful in dealing with more general situations. Technically speaking, $g(x_0)$  and $Dg(x_0)$ can be absorbed into other parameters. Here we choose to display them to track the dependence of relevant parameters on the dimension $n$.

\begin{theo}\label{linearcontrol1}Suppose that $\bar p=\bar p(t,x)$ is the solution to heat equation \eqref{heat}.
Let   $|v|_{1}=\max\{|v|,1\}$ for $v\in \Rset^n$.  Fix $x_0\in \Rset^n$.

(i)  Given Assumption \ref{assp:max-logconvx} and   $|\nabla g(x)|\leq \beta_1|x-x_0|+\beta_2$ for  $\beta_1,\beta_2\geq 0$.  Then for all $(t,x)\in [0,1]\times \Rset^n $,
\be\label{1storderestimate}
|\nabla\bar q(t,x)|\leq {3\beta_1\over \sqrt{1-\alpha_2}}\max\{C_n, \ C_{\beta_1,\alpha_2}|x-x_0|_1\}+\beta_2.
\ee

Here the two constants $C_n=2\sqrt{(n+3)\log \left({2(1+4\beta_1)\over \sqrt{1-\alpha_2}}\right)+4n\log n+\alpha_1+1+{{{\beta_2}^2\over \beta_1}}}$ and $C_{\beta_1,\alpha_2}=3\sqrt{\beta_1}+1+{6\alpha_2\over \sqrt{1-\alpha_2}}$.

(ii) Assume $||D^2g(x)||_2\leq L$ and there exists $x_0$, $\alpha_2\in[0,1)$\footnote{In particular, if $g$ attains minimum at some point $x_0$, then we can take $\alpha_2=\alpha_1=0$.}, such that
\begin{align}\label{assp:max-logconvx-v2}
    g(x)-g(x_0)-\nabla g(x_0)\cdot (x-x_0)\geq -{\alpha_2\over 2}|x-x_0|^2-\alpha_1 \quad \forall x\in\Rset^n .
\end{align}  Then for  all $(t,x)\in [0,1]\times \Rset^n $, \

\be\label{2ndorderestimate}
||D^2\bar q(t,x)||\leq  {10L^2+L\over 1-\alpha_2}\max\left\{\tilde C_n^2, \   (\tilde C_{L, \alpha_2})^2(|x-x_0-\nabla g(x_0)|_1)^2\right\} ,
\ee
\be\label{testimate}
|\nabla \bar q_t(t,x)|\leq   {48L^2+2L\over \sqrt{t}(1-\alpha_2)^{3\over 2}}\max\left\{\tilde C_n^{3}, \  (\tilde C_{L, \alpha_2})^3(|x-x_0-\nabla g(0)|_1)^3\right\}.
\ee
Here the two constants $\tilde C_n=2\sqrt{(n+3)\log \left({2(1+4L)\over \sqrt{1-\alpha_2}}\right)+4n\log n+\alpha_1+1}$ and $\tilde C_{L, \alpha_2}=3\sqrt{L}+1+{6\alpha_2\over \sqrt{1-\alpha_2}}$.

\end{theo}

The proof is in Section \ref{proof:linearcontrol1}. A simpler case with bounded $\nabla g$ is also discussed. See Remark \ref{moreont} for another way to bound $|\nabla \bar q_t(t,x)|$ by replacing $L^2\over \sqrt{t}$ in  \eqref{testimate} by $O(nL^3)$.

As an immediate corollary,  we have 
\begin{cor}\label{cor:n-control}  Under the same assumption as (ii) in Theorem \ref{linearcontrol1},
we further assume there exists $C_0>0$ such that,  $\alpha_1\leq C_0n$ and $|\nabla g(x_0)|\leq C\sqrt{n}$.
Then
$$
\begin{array}{ll}
&||D^2\bar q(t,x)||\leq CL^2\left(n\log n+L|x-x_0-\nabla g(x_0)|^2\right)\\
&|\nabla \bar q_t(t,x)|\leq CL^2\left({(n\log n)}^{3\over 2}+L\sqrt{L}|x-x_0-\nabla g(x_0)|^3\right).
\end{array}
$$
Here $C$ is a constant independent of $L$ and $n$. 
\end{cor}
Note that the further assumptions of scale relates to normalization in $n$  dimension.

\begin{rmk}\label{n-control} Owing to \ref{all-in-one} in the proof of Theorem \ref{linearcontrol1}),  under the assumption of corollary \ref{cor:n-control}, we have for all $m\in \Nset$
$$
\mathbb{E}_{p(t,x)}(|x(t)|^m)\leq O\left(L^{m\over 2}(n\log n)^{m\over 2}\right)
$$
This demonstrates that,  if we only care about expectations of powers of $D^2\bar q(t,x)$ or $\nabla \bar q_t(t,x)$,   $\|D^2\bar q(t,x)\|_2$ behaves like $O(L^3n\log n)$ and  $|\nabla \bar q_t(t,x)|$ behaves like $O\left(L^3\sqrt{L}(n\log n)^{3\over 2}\right)$. Note that the point $x_0$ itself plays no role in computing the expectation that is translation invariant in the $x$ variable.

\end{rmk}

\begin{theo}\label{main:theo1}  Let $g(x)\in C^{0,1}(\Rset^n)$ satisfy the Assumption \ref{assp:max-logconvx} and $|\nabla g|\leq C(|x|+1)$ for a positive constant $C$. Then for any $T>0$,   the above \eqref{mainode} is well-posed.

\end{theo}

Proof:  Note that $q(t,x)$ is a smooth function, hence locally Lipschitz continuous in $x$. Owing to Theorem \ref{well-posed}, it suffices to show that  for $q=-\log p(t,x)$,
$$
|Dq(t,x)|\leq C_T(|x|+1) \quad \text{for all $(t,x)\in  [0,T]\times \Rset^n$}
$$
for a constant $C_T$ depending on $C$ and $T$. By \eqref{transform}, it
is equivalent to showing that 
$$
|D\bar q(t,x)|\leq C_T(|x|+1) \quad \text{for all $(t,x)\in  [0,1-e^{-T}]\times \Rset^n$}.
$$
for a constant $C_T$ depending on $C$ and $T$,  which follows from Theorem \ref{linearcontrol1}. \qed
\subsection{Compactly Supported Data Distributions}\label{sec:compact}

In this section,  we look at the situation where the data distribution $p_0$ is a positive measure with compact support which is a typical situation in image generation \cite{B2023}. 
Due to the manifold hypothesis, the support is typically a low dimension set.  In this situation,  what is important is the asymptotic estimate as $t\to 0$.  The following are two estimates 
related to \cite{B2023}. 

\begin{theo}\label{linearcontrol2} We assume $\mathrm{supp}(p_0)=D_0\subset \overline {B_M(0)}$ and the density is smoothly defined on $D_0$. We have,
\begin{flalign}
   &\text{(1)}  \quad 
|\nabla\bar q(t,x)|\leq {|x|+M\over t}; \\
&\text{(2)} \quad
||D^2\bar q(t,x)||_2\leq {1\over t}+{M^2\over t^2}\label{l-bound}.
\end{flalign}
 \end{theo}
 The proof is in Section \ref{proof:linearcontrol2}

Fixing $x$, denote by $\bar y_t$  the weighted center of mass:
$
\bar y_t={\int_{D_0}ye^{-|x-y|^2\over 2t}d\pi_0(y)\over \hat p}
$.
For a ``regular"  $\pi_0$,  as $t\to 0$,  we expect the measure $e^{-|x-y|^2\over 2t}d\pi_0(y)\over \hat p$ will concentrate on $\{y\in D_0|\ |y-x|=d(x, D_0)\}$.  Thus, 
$$
\lim_{t\to 0}d(\bar y_t, \Gamma_x)=0,
$$
where $\Gamma_x$ is the convex hull of $\{y\in D_0|\ |y-x|=d(x, D_0)\}$.  Then
$$
|\nabla\bar q(t,x)|={|x-\bar y_t|\over  t} \quad \mathrm{and} \quad \liminf_{t\to 0}t|\nabla\bar q(t,x)|\geq d(x,\Gamma_x).
$$
So if $x\notin \Gamma_x$ (typical situation for low dimension set $D_0$), then $|D\bar q(t,x)|=O({1/ t})$. Hence the $1/ t$ blow up for the gradient bound is usually inevitable, which matches 
experimental observations \cite{score_exploding_2022}.

Therefore, in practice, the initial distribution that we are really getting  from the denoising process is $p(t_0,x)$ for a small $t_0>0$. Per previous discussions, it is important to have a better global Lipschitz bound of $\log p(t_0,x)$ than $O(1/ t^2)$.  As a first step toward this direction,   the following theorem says that the typical Hessian bound is actually $O\left(1/ t\right)$ instead of  $O\left(1/ t^2\right)$ with the exception of a small set.  Accordingly, in practice, it might be reasonable to assume that the Hessian is bounded by $O({1/ t})$.   For clarity of presentation,  we will assume that $D_0$ is a low dimensional smooth manifold with boundary although our results can be properly extended to manifolds with less regularity.   An example will be presented to show that our conclusion is optimal, see Example \ref{eg:Hessian-growth-t2}. 

\begin{theo}\label{theo:optimalbound} For $1\leq d\leq n$, assume that $D_0\subset \Rset^n$ is a $d$-dimensional compact smooth manifold with boundary  and $\pi_0$ is comparable to the uniform distribution on $D_0$.  Then for almost everywhere $x\in \Rset^n$, 
$$
||D^2\bar q(t,x)||_2\leq {C_x\over t} \quad \text{for $t\in [0,1]$}.
$$
Here $C_x$ is a constant depending only on $x$ and $D_0$.  If  $D_0$ is convex, then the above holds for all $x\in \Rset^n$. 
\end{theo}
Proof is in Section \ref{prof:theo:optimalbound}

We will present a smooth non-convex $D_0$ that shows the result of Theorem \ref{theo:optimalbound}  is optimal. 
\begin{example}\label{eg:Hessian-growth-t2} 
Let $D_0\subset \Rset^2$ be the domain obtained by removing a small square $[0,2]\times [-1,1]$ from the big square $[-2,2]^2$ and then mollifying the corners to make it smooth. Here $O=(0,0)$.  The $Y$-shaped region 
$$
L=\{x\in \Rset^2|\ \text{there are more than one $y$ such that $|x-y|=d(x,D_0)$}\}.
$$
We also choose $\pi_0={\chi_{D_0}\over |D_0|}dx$, i.e., the uniform distribution on $D_0$, where $|D_0|$ is the area of $D_0$.
\begin{center}
\includegraphics[scale=0.8]{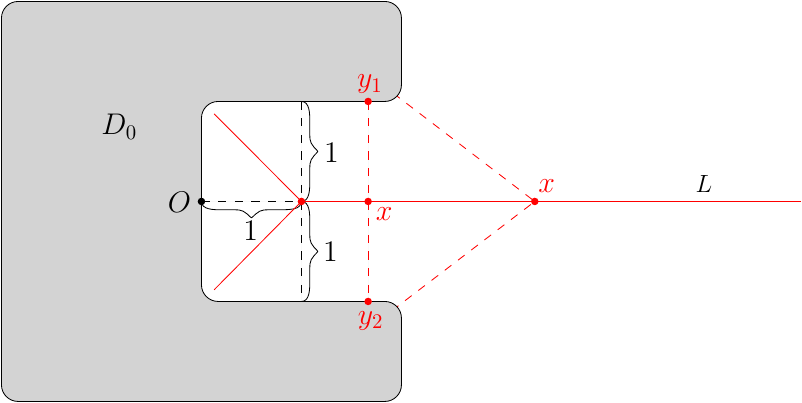} \quad \qquad
\captionof{figure}{In the above picture, $\Gamma_x=\{sy_1+(1-s)y_2: s\in [0,1]\}$}
\end{center}

We have that,
$
||D^2\bar q(x,t)||_2\geq {C_x/t^2}   \quad \text{for $x\in L$ and $t\in (0,1]$}$.
\end{example}
For reader's convenience, we will verify  the above when $x=(\theta,0)$ for $\theta>1$ in Section \ref{proof:eg:Hessian-growth-t2}. The other parts are left to interested readers as an exercise. 

\section{Well-posedness and Convergence under Sharp Lipschitz Bound}\label{sec:convergence}

As the starting point, we provide a well-posedness condition of a general SDE with additive noise.
 
\begin{theo}\label{well-posed} Given $T>0$,   suppose that $F=F(t,x)\in C([0,T]\times \Rset^n, \Rset^n)$ satisfies that $F$ is locally Lipschitz continuous in $x$ variable, i.e., for any $M>0$, there exists a constant $L_M$ such that
$$
|F(t,x)-F(t,y)|\leq L_M|x-y| \quad \text{for $x,y\in B_M(0)$ and $t\in [0,T]$}
$$
and
\be\label{growth}
|F(t,x)|\leq C(|x|+1).  \quad \text{for $(t,x)\in  [0,T]\times \Rset^n$}.
\ee
for a positive constant C.  For any $x_0\in \Rset^n$, the following SDE has a unique solution
$$
dX_t=F(t,X_t)dt+dW_t, \quad \text{$t\in [0,T]$}, \;\; 
X_0=x_0.
$$
\end{theo}
The proof is in Section \ref{proof:well-posed}.
\begin{rmk}
Note that the uniform Lipschitz continuity of $F$ is not needed given the linear growth condition \eqref{growth}, which is  known to experts. See Theorem 2.4 and Theorem 3.1 in Chapter IV of \cite{ikeda2014stochastic} for more general situations.  For reader's convenience,  we presented the proof above for our special case.
    \end{rmk}
    
Due to the limitation of global Hessian estimate, the convergence analysis is divided into the following two cases by $p_0$. The first case enjoys better complexity with respect to dimension ($N=\mathcal{O}(n)$) while has limitation in the final time $T$ in the forward process. The second case is valid globally in $T$ while  achieving polynomial complexity ($N=\mathcal{O}(n^3\log^2 n)$). 

\paragraph{Case I: $p_0$ is (near) log-concave}

\begin{theo}\label{thm:conv-0}
      Assume the following global Hessian bound of $p_0$, $$-L_1 I_n \preceq D^2\log p_0(x)\preceq L_0 I_n.$$ Let $\hat{q}_{T}$ be a distribution generated by the uniform discretization ($\delta=0$) of the exponential integrator scheme \eqref{semi-discrete}, with an approximated score satisfying the Assumption \ref{ass:score}.
     For $L_1> 0$,
    \begin{align*}
\mathrm{KL}(p_0\| \hat{q}_{T}) \lesssim (M_2+d)e^{-T}+T\epsilon_0^2+\frac{dT^2H_T^2}{N},
\end{align*}
where $H_T$ depends on $L_0$, $L_1$, $T$ and $T<-\log(1-\frac{1}{L_0+1})$. 

If $L_1\leq 0$, namely $p_0$ is log-concave,
\begin{align*}
\mathrm{KL}(p_0\| \hat{q}_{T}) \lesssim (M_2+n)e^{-T}+T\epsilon_0^2+\frac{nT^2H^2}{N},
\end{align*}
where the constant $H$ depends on $L_0$ and $L_1$.
\end{theo}
\textbf{Proof:} We first apply Corollary\ref{Hessianbound} to attain global Hessian estimate in finite time. Then apply it to Theorem \ref{well-posed} for well-posedness and Proposition \ref{est-global-lip} for convergence rate.

\paragraph{Case II: General smooth $p_0$} 
\begin{theo}\label{thm:conv-1}
    With the same assumption of Corollary \ref{cor:n-control}, let $\hat{q}_{T}$ be a distribution generated by  uniform discretization of the exponential integrator scheme \eqref{semi-discrete}, with an approximated score satisfying Assumption \ref{ass:score}. Then we have,
    \begin{align}
    \mathrm{KL}(p_0\| \hat{q}_{T}) \lesssim (M_2+d)e^{-T}+T\epsilon_0^2+{CL^6Tn(n\log n)^2\over N}.
    \end{align}
\end{theo}
\textbf{Proof:} This is a direct consequence of Theorem \ref{thm:accumuated-error} to estimate truncation error in Proposition \ref{prop:error-decomp}.

In addition to the above cases, we also consider non-smooth $p_0$ supported on compact manifold. Restricted by the estimate in Theorem \ref{linearcontrol2}, we switch to the early stopping technique, namely $\delta>0$ in discretization. Due to the measure zero set (see in Section \ref{sec:compact}), the convergence bound is not yet optimal as shown in Section \ref{sec:conv-compact}.

\section{Conclusion}
In this paper, we analyzed the Lipschitz bounds of the score in the SGM. Our bounds are sharp in light of the constructed counter-examples. Based on the result, we provide the guarantees for SGM in the framework where $L_2$ accurate score estimator is available and smoothness assumption holds on the data distribution. Our bounds for the non-smooth case characterize singular behaviours of the score near the generation time, offering insights for model parameterization in practice. 

\paragraph{Limitation}
Due to the limited knowledge of regularity factors of data distribution (e.g. optimal Lipschitz constants), our bound cannot provide implementable guidance on seeking the optimal schedule (which may require a separation of temporal regimes). Also in the manifold case, due to the complex geometries, as shown in the non-smooth section, our theories cannot provide a justifiable guidance of early stopping time. We will investigate these issues in a future study.

\appendix
\section{Convergence theories}
In this section, we list some numerical algorithms and convergence theories related to the numerical discretization of \eqref{mainode}. They are due to \cite{chen2023improved}.

\paragraph{Convergence in distribution } The key ingredient of the convergence theory is the following result from the chain rule of KL divergence.
\begin{prop}[Prop C.3. of \cite{chen2023improved}]\label{prop:error-decomp}
    Given the score error estimation Assumption    \ref{ass:score}, the exponential integrator scheme \eqref{semi-discrete} satisfies,
\begin{align*}
\mathrm{KL}(p_\delta\| \hat{q}_{T-\delta}) \lesssim \mathrm{KL}(p_T\| \gamma_n) + T\epsilon_0^2 + \sum_{k=1}^N \int_{t_{k-1}}^{t_k} E \|\nabla \log p_t(x_t) - \nabla \log p_{t_k}(x_{t_k}) \|^2 \d t,
\end{align*}
where $\gamma_n$ is the Uniform Gaussian distribution
\end{prop}
The first term $\mathrm{KL}(p_T\| \gamma_n)$ in the estimate measures the distance between  the measure of OU process to its invariant measure. When the data has finite second order moment, it tends to $0$ as $T\to \infty$.
\begin{prop}[Lem C.4 of \cite{chen2023improved}]With finite second order moment $E_{p_0}|X|^2<\infty$, for $T>1$,
\begin{align*}
    KL(p_T,\gamma_n)\leq (n+M_2)e^{-T}.
\end{align*}
\end{prop}

The third term relates to local truncation error that depends on the regularity of the forward process. Then Proposition  \ref{prop:error-decomp} can be further extended if the global Hessian estimate is available.
\begin{prop}[Theorem   2.1 of \cite{chen2023improved}]\label{est-global-lip}
    Given assumption of Proposition  \ref{prop:error-decomp} , $\nabla \log p_t$ is $L$-Lipschitz. For uniform discretization, the exponential integrator scheme \eqref{semi-discrete} satisfies,
    \begin{align*}
\mathrm{KL}(p_0\| \hat{q}_{T}) \lesssim (M_2+n)e^{-T}+T\epsilon_0^2+\frac{nT^2L^2}{N}.
\end{align*}
\end{prop}
\section{Semi-convexity of second order differential equation}
Here we list some important theories for constructing the finite time log-convexity of the density of forward process.

 Given a function $w(t,x)$, its convex envelop $w^{**}(t,x)$  in $x$ is defined as 
\begin{align}\label{defi-convex}
     w^{**}(t,x)=\inf\left\{\sum_{i=1}^{n+1}\lambda _iw(t,z_i)|\ x=\sum_{i=1}^{n+1}\lambda_iz_i,\ \sum_{i=1}^{n+1}\lambda_i=1,\ \lambda_i\geq 0, z_i\in \Rset^n\right\}.
\end{align}

\begin{lem}[Prop 7 of \cite{ALL1997}, Lemma 2 of \cite{S2010}]\label{ALL-convex}
Let $w$ be a solution of 
\begin{align}\label{eqn-w-general}
    w_t +F(t,x,\nabla w,D^2 w) =0.
\end{align}
The convex envelope $w^{**}$ of $w$ is a supersolution of \eqref{eqn-w-general}, under the following assumptions,
\begin{enumerate}
\item $F$ is elliptic in the sense $F(t,x,p,A)\geq F(t,x,p,\tilde A)$ if $A\leq \tilde A$.
\item $(x,A)\in \Rset^n\times S_{++}^{n}\mapsto F(t,x,p,A^{-1})$ concave for all $t$ and $p$. Here $S_{++}^{n}$ is the set of all $n\times n$ positive definite matrices.
\item $w$ is coercive in the sense that 
\begin{align}\label{coercivity-assumption}
\lim_{x\to\infty} \frac{w(t,x)}{|x|}=\infty,
\end{align}
uniformly in $t$.
\end{enumerate}
\end{lem}

\section{Proofs}
In this section, we collects the proofs to the results.

\subsection{ Proof of Theorem \ref{secondestimate}}\label{proof:secondestimate}  For (1), it is equivalent to showing that
$$
D_x^2\bar q(t,x)\leq  M_1I_n \quad \text{for all $(t,x)\in [0,1]\times \Rset^n$}.
$$
This actually follows immediate from \eqref{Hessianexpression}. Here we will present a standard PDE approach that does not rely on the formula.

Let $\xi$ be a given unit vector.  By taking derivatives of \eqref{vHJ}, we deduce that $v=q_{\xi\xi}$ satisfies
$$
v_t-{\frac{1}{2}}\Delta v+\nabla\bar q\cdot \nabla v=-|\nabla\bar q_{\xi}|^2\leq 0  \quad \text{on $(0,1)\times \Rset^n$}.
$$
Thanks to  the standard maximum principle of parabolic equation,  we have that
$$
v(t,x)\leq \sup_{x\in \Rset^n}v(x,0)=\sup_{x\in \Rset^2}q_{\xi\xi}\leq M_1. 
$$

The proof of (2) is more interesting. It is equivalent to showing that 
\be\label{shortimeforbarq}
D^2\bar q(t,x)\geq -\frac{M_0}{1-M_0t} I_n.
\ee
Below we show a PDE approach based on a modification of the arguments in \cite{S2010} for obtaining semiconcavity of solutions to the general viscous Hamilton-Jacobi equations.  

Fix $\delta_1>0$.  Note that by $D^2g\succeq -M_0I_n$,
$$
g(x)-g(0)-\nabla g(0)\cdot x\geq -{M_0\over 2}|x|^2.
$$
Hence there exists a constant $C_{\delta_1}$ such that
$$
g(x)\geq -\left({M_0\over 2}+{\delta_1}\right)|x|^2-C_{\delta_1}.
$$
 Let $\alpha$ and $c$ be positive numbers satisfying  
 \be\label{setT}
 \theta(0)=\alpha \tan(\alpha c) \geq M_{\delta_1}={M_0}+4{\delta_1}
   \ee
Consider the following construction,
\begin{align}\label{defi-w}
    w = \bar q + \theta(t)|x|^2/2 +n\Theta(t)/2,\\
\end{align}where,
\begin{align}
    &\theta(t)=\alpha \tan(\alpha c +\alpha t), \quad t< T^* =\frac{\pi}{2\alpha}-c  \label{eq:gt} \\
    &\Theta(t)=\int_0^t \theta(s)ds.
 \end{align}

We notice that Eq.\eqref{eq:gt} implies, $\theta(0)=\alpha \tan(\alpha c)$ and $\theta'-\theta^2=\alpha^2$.
Then $w$ satisfies the following equation:
\begin{align}
0=w_t +F(t,x,\nabla w, \nabla^2 w):=w_t -\frac{1}{2} \Delta w +\frac{1}{2}|\nabla w|^2 -\theta(t)\nabla w\cdot x -\frac{\alpha^2}{2} |x|^2\label{eqn-w}.
 \end{align}

 Now we consider the convex envelope (definition see \eqref{defi-convex}) of $w$, $w^{**}$ and aim to apply Lemma    \ref{ALL-convex} to show that $w^{**}$ is a supersolution of Eq.\eqref{eqn-w}. After direct validation of the first two conditions of Lemma   \ref{ALL-convex}, it resorts to coercivity assumption \eqref{coercivity-assumption}. To this end, we construct a solution $\underline{q}$ of the equation (\eqref{vHJ}) subjecting to $\underline q(0,x)\leq \bar g(x)$:
 \begin{equation}
     \underline{q}=\theta_1(t)\frac{|x|^2}{2} +\Theta_1(t)\frac{n}{2}-C_{\delta_1}
 \end{equation}
 where
 \begin{equation}
 \label{cond-underq}
          \theta_1(t)=\frac{1}{t-\frac{1}{M_0+2{\delta_1}}}, \quad 
          \Theta_1(t)=\int_0^t\theta_1(t)dt.
 \end{equation}
 From Eq.\eqref{cond-underq},  we know the construction holds for $t\in[0,{1\over M_0+2{\delta_1}})$.
 By revisiting \eqref{eq:gt},
 \be\label{maxT}
   \sup_{\{\alpha \tan(\alpha c) \geq M_{\delta_1}\} }\left(\frac{\pi}{2\alpha}-c\right)= \lim_{\{\alpha\to 0^+,\ \alpha \tan(\alpha c)=M_{\delta_1}\} }\left(\frac{\pi}{2\alpha}-c\right)={1\over M_{\delta_1}}={1\over M+4{\delta_1}}
    \ee
   and
    \begin{align}\label{optimal:lowerD2}
      \lim_{\{\alpha\to0^{+}, \ \alpha \tan(\alpha c)=M_{\delta_1}\}}\theta(t)=\frac{M_{\delta_1}}{1-M_{\delta_1} t}.
    \end{align}
    
 Then by choosing suitable $\alpha$ and $c$, the comparison principle of \eqref{vHJ} which is equivalent to the one for  \eqref{heat} says that
 $$
 \bar q(x,t)\geq \underline{q}(x,t)\quad \text{for $t\in [0, {1\over M_{\delta_1}})$}
 $$
 and hence, 
 \begin{align}\label{coer-w}
     w\geq (\theta(t)+\theta_1(t))\frac{|x|^2}{2} +(\Theta(t)+\Theta_1(t))\frac{n}{2}\quad \text{for $t\in [0, {1\over M_{\delta_1}})$}.
 \end{align}

 Now turning back to Eq.\eqref{coer-w}, we know $\theta(t)+\theta_1(t)\geq 2{\delta_1}>0$ uniformly in any closed subinterval of  $t\in [0,{1\over M_{\delta_1}})$. This ensures the uniform coercivity requirement in Lemma   \ref{ALL-convex}.
  
Summing up, by Lemma   \ref{ALL-convex}, $w^{**}$ is a supersolution. On the other side, as convex envelope, $w^{**}\leq w$. Next, we want to utilize the comparison principle of  \eqref{eqn-w} to show $w^{**}\geq w$ for all $t$, which is equivalent to Eq.\eqref{vHJ} due to the construction \eqref{defi-w}. To do this, we only need $w^{**}(0,X)\geq w(0,X)$, equivalently $w(0,X)$ is convex and is assured by $\theta(0)\geq M_{\delta_1}>M_0$.

Now we have $w=w^{**}$ for $x\in R^d$, $t\in [0,T^*]$, is convex. In particular, this implies that 
$$
D^2\bar q(t,x)\geq -\theta(t)I_n  \quad \text{for $(t,x)\in [0,T^*]\times \Rset^n$}.
$$

Hence we derive that for any $T<{1\over M_{\delta_1}}$, 
\begin{align}\label{est:semi-convex1}
    \inf_{[0,T]\times \Rset^n}D^2\bar q(t,x)\geq -\frac{M_{\delta_1}}{1-M_{\delta_1} t} I_n.
\end{align}

Then \eqref{shortimeforbarq} follows by sending ${\delta_1}\to 0$.

Since the transform \eqref{transform} only requires estimate of $\bar q$ for $t\in [0,1]$, when $M_0<1$, \eqref{est:semi-convex1} holds for any $T<1$. Recalling transformation of $q$, the condition $M_0<1$ is equivalent to $-\log p$ is convex.

 \qed

\subsection{Construction of Example of the Loss of Uniform Hessian Bound in Section \ref{eg:counter-eg}}\label{proof:counter-eg}

Precisely speaking,  we will construct a one dimensional ($n=1$) example  of $g(x)=G^2(x)$ for a smooth Lipschitz continuous function $G$  satisfying that $|G'(x)|\leq 1$,   $|g''(x)|\leq 2=M_0=M_1$ and 
$$
\limsup_{|x|\to +\infty}\left |\bar q''\left(1/2,\ x\right)\right|=\infty.
$$
For $M > 0$,  let $g_M$ be the even function such that
$$g_M(x) = \begin{cases}
2M^2 - x^2, \quad 0 \leq x \leq M \\
(x-2M)^2, \quad M \leq x \leq 2M \\
0, \quad x \geq 2M.
\end{cases}$$
Note that $|g_M''| \leq 2$ independent of $M$. Let $h_M(t,\ x)$ be the solution to the following heat equation
\begin{align}\label{heateq}
   u_t-{\frac{1}{2}}\Delta u=0\quad \text{for $(t,x)\in (0,\infty)\times \Rset$} 
\end{align}
subject to $h_{M}(0,x)=e^{-g_M(x)}$. 

Since $h_M$ is even in $x$, $(h_M)_x(1/2, \ 0)=0$.  Hence, we have
\begin{align*}
&(\log h_M)_{xx}(1/2,\ 0) = \frac{(h_M)_{xx}(1/2,\ 0)}{h_M(1/2,\ 0)} \\
&= \frac{e^{-2M^2}\int_0^M (4y^2+2)\,dy + e^{-2M^2} \int_{M}^{2M} e^{-2(y-M)^2}(4(y-2M)^2-2)\,dy}{e^{-2M^2}M + e^{-2M^2}\int_{M}^{2M} e^{-2(y-M)^2}\,dy + \int_{2M}^{\infty} e^{-y^2}\,dy} \\
&:= \frac{A + B}{C + D + E}.
\end{align*}
 Clearly, for $M\geq 1$, 
$$
A+B=4e^{-2M^2}\int_0^M y^2\,dy + 4e^{-2M^2} \int_{M}^{2M} e^{-2(y-M)^2}(y-2M)^2\,dy>M^3e^{-2M^2}.
$$
and
$$
C+D+E\leq 3Me^{-2M^2}.
$$
Thus
\begin{equation}\label{blockbound}
(\log h_M)_{xx}(1/2, 0) > {M^2\over 3}
\end{equation}
for all $M\geq 1$.
\begin{rmk}
Note that $g_M$ is $C^{1,1},$ not smooth. However, the estimates above still hold  for sufficiently fine mollifications of $g_M$, so we may assume without loss of generality that $g_M$ is smooth.
\end{rmk}

Below we will choose a sequence $0 \leq x_1 \leq x_2 \leq ...$ such that the terms in $g := \sum_{k = 1}^{\infty} g_k(x - x_k)$ have disjoint support,  and such that  
\be\label{lowerbound}
(\log \bar p_N)_{xx}(1/2,\ x_k) > {k^2\over 3},\quad \text{for $k = 1,\,...,\,N$},
\ee
where $\bar p_N$ is the solution to \eqref{heateq} subject to $\bar p_{N}(0,x)=e^{-\sum_{k = 1}^N g_k(x - x_k)}$. Note that $\bar p_{N}(0,x)=1$ for $x\leq -2$. 

Suppose we have managed to do this. Note that
$$
\hat p\leq \bar p_N\leq 1
$$
where $\hat p$ is the solution to \eqref{heateq} subject to $\hat p(0,x)=\chi_{[-3,-2]}$. Then interior derivative estimates for solutions of \eqref{heateq}  imply that $(\log \bar p_N)_{xx}(1/2, \ \cdot)$ converges locally uniformly on $\mathbb{R}$ as $N \rightarrow \infty$ to $(\log \bar p)_{xx}(1/2, \ \cdot,)$, where $q$ is the caloric function with initial data $e^{-g}$. Therefore, $(\log \bar p)_{xx}(\cdot,\, 0)$ is bounded, but $(\log \bar p)_{xx}(1/2,\ \cdot)$ is unbounded (its values at $x_k$ are at least $k^2\over 3$), as desired. In addition, it is easy to check $\int_\Rset p(0,x) dx =\int_\Rset \exp(-g(x)-\frac{x^2}{2})dx<\int_\Rset \exp(-\frac{x^2}{2}) dx <\infty$. This implies the constructed $g$ corresponds to a density function after the transform \eqref{transform} and appropriate normalization.

We now explain how to choose $x_k$. We will repeatedly use the fact that if $h$ is a bounded smooth function on $\mathbb{R}$ and $\tilde{h}$ is a compactly supported smooth function, then the caloric function with initial data $h + \tilde{h} (\cdot + S)$ converges in $C^2$ as $|S| \rightarrow \infty$ on compact subsets of $\{t > 0\}$ to the caloric function with initial data $h$. 

First, we let $x_1 = 0$. Then \eqref{lowerbound} with $N = 1$ follows immediately from \eqref{blockbound}. Now suppose we have chosen $x_1 < ... < x_{M-1}$  such that the supports of $g_k(x -x_k),\, 1 \leq k \leq M-1$ are disjoint and \eqref{lowerbound} holds for $N = M-1$. Using the above-mentioned fact and \eqref{blockbound}, if we take $x_{M}$ sufficiently large, then \eqref{lowerbound} holds for $N = M$. Indeed, the inequality for $k < M$ follows immediately from the fact above, and the inequality for $k = M$ follows from the fact above and the inequality \eqref{blockbound}, after translating so that $x_M$ becomes $0$. This completes the construction.
\subsection{Proof of Theorem \ref{linearcontrol1}}\label{proof:linearcontrol1} Without loss of generality, we may assume $x_0=0$.  It suffices to show the above for $|x|\geq 1$.  For $|x|\leq 1$, we can just replace $|x|$ in all the final bounds with 1.

First, we prove \eqref{1storderestimate}.  Without loss of generality, let $g(0)=0$. Then
$$
g(z)\leq {\beta_1\over 2}|z|^2+\beta_2|z| \quad \text{for $z\in \Rset^n$}. 
$$
Recall that
\be
\begin{array}{ll}
\bar p(t,x)&={1\over {(2\pi t)}^{n\over 2}}\int_{\Rset^n}e^{-{|x-y|^2\over 2t}}h(y)\,dy\\[5mm]
&={1\over {(\pi )}^{n\over 2}}\int_{\Rset^n}e^{-{|y|^2}}h(x-\sqrt{2t}y)\,dy.
\end{array}
\ee
Then
\be
\begin{array}{ll}\label{gradientexpression}
-\nabla \bar q={\nabla \bar p\over \bar p}&={1\over \bar p}{1\over {(\pi )}^{n\over 2}}\int_{\Rset^n}e^{-{|y|^2}}h(x-\sqrt{2t}y)Dg\,dy\\[5mm]
&={1\over \bar p}{1\over {(\pi )}^{n\over 2}}\int_{\Rset^n}e^{-{|y|^2}}h(x-\sqrt{2t}y)Dg\,dy
\end{array}
\ee
Since $ab\leq a^2+{b^2\over 2}$, 
$$
g(x-\sqrt{2t}y)\leq \beta_1(|x|^2+2|y|^2)+\beta_2(|x|+\sqrt{2}|y|)\leq 2\beta_1(|x|^2+2|y|^2)+{{\beta_2}^2\over \beta_1},
$$
we deduce that
\be\label{eq4}
\begin{array}{ll}
\bar p(x,t)&\geq e^{-{{\beta_2}^2\over \beta_1}}{e^{-2\beta_1|x|^2}\over {(\pi )}^{n\over 2}} \int_{\Rset^n}e^{-(1+4\beta_1){|y|^2}}\,dy\\[5mm]
&=e^{-{{\beta_2}^2\over \beta_1}}{e^{-2\beta_1|x|^2}} ({1\over 1+4\beta_1})^{n\over 2}.
\end{array}
\ee

Then
$$
\begin{array}{ll}
|\nabla\bar p|&={1\over {(\pi )}^{n\over 2}}\left|\int_{\Rset^n}e^{-{|y|^2}}Dh(x-\sqrt{2t}y)\,dy\right|={1\over {(\pi )}^{n\over 2}}\left|\int_{\Rset^n}e^{-{|y|^2}}hDg(x-\sqrt{2t}y)\,dy\right |\\[5mm]
&\leq {1\over {(\pi )}^{n\over 2}}\int_{\Rset^n}(\beta_1|x|+\sqrt{2}\beta_1|y|+\beta_2)e^{-{|y|^2}}h(x-\sqrt{2t}y)\,dy\\[5mm]
&=(\beta_1|x|+\beta_2)\bar p+{\sqrt{2}\beta_1\over {(\pi )}^{n\over 2}}\int_{\Rset^n}|y|e^{-{|y|^2}}h(x-\sqrt{2t}y)\,dy.
\end{array}
$$

Let
$$
\tilde  K={2\over \sqrt{1-\alpha_2}}\max\left\{{C_{n,m}\over |x|} , \ 4\sqrt{\beta_1}+1\right\},
$$
where
$$
C_{n,m}={2} \sqrt{(n+3)\log \left({2(1+4\beta_1)\over \sqrt{1-\alpha_2}}\right)+4n\log n+\alpha_1+J_m+{{{\beta_2}^2\over \beta_1}}}.
$$
Here for $m\in \Nset$, $J_m$ is the last positive integer such that $e^{r^2}\geq r^m$ for when $r\geq J_m$.  In particular, $J_1=J_2=J_3=1$ and $C_n=C_{n,m}$ for $m=1,2,3$.

\medskip

{\bf Claim:} If 
$$
K\geq K_0= \max\left\{\tilde K, \ {6\alpha_2\over 1-\alpha_2}\right\},
$$
then for $i=1,2,3,...,m$, 
\be\label{all-in-one}
T_i(x)={1\over \bar p(t,x)}{1\over {(\pi )}^{n\over 2}}\int_{y\in \Rset^n}|y|^{i}e^{-{|y|^2}}h(x-\sqrt{2t}y)\,dy\leq K^i|x|^i+1 \leq 2K^i|x|^i.
\ee
Clearly,  $T_i(0)=\mathbb{E}_{p(t,z)}(|z|^i)$.

Let us prove the claim. Note that
$$
\begin{array}{ll}
\bar p(t,x)T_i &={1\over {(\pi )}^{n\over 2}}\int_{|y|\leq K |x|}|y|^ie^{-{|y|^2}}h(x-\sqrt{2t}y)\,dy+{1\over {(\pi )}^{n\over 2}}\int_{|y|\geq K|x|}|y|^ie^{-{|y|^2}}h(x-\sqrt{2t}y)\,dy\\[5mm]
&\leq  K^i|x|^i\bar p(x,t)+\underbrace{{1\over {(\pi )}^{n\over 2}}\int_{|y|^i\geq K|x|}|y|^ie^{-{|y|^2}}h(x-\sqrt{2t}y)\,dy}_{\mathrm{I_i}}.
\end{array}
$$

Our goal is to show that  $I_i\leq \bar p(t,x)$ when $K\geq K_0$. Since $g(z)\geq -\alpha_2|z|^2-\alpha_1$, $|y|\geq K|x|$ and $K>{6\alpha_2\over 1-\alpha_2}$,
$$
\begin{array}{ll}
g(x-\sqrt{2t}y)&\geq -{\alpha_2\over 2}\left(|x|+\sqrt{2}|y|\right)^2-\alpha_1\\[5mm]
&> -\alpha_2|y|^2\left({1\over K}+1\right)^2-\alpha_1\\[5mm]
&> \alpha_2|y|^2\left({3\over K}+1\right)-\alpha_1\\[5mm]
&\geq -{(1+\alpha_2)\over 2}|y|^2-\alpha_1
\end{array}
$$
Then
$$
\mathrm{I_i}\leq {e^{\alpha_1}\over (\sqrt{\pi})^n}\int_{|y|\geq K|x|}|y|^ie^{-{(1-\alpha_2)|y|^2\over 2}}\,dy.
$$

For convenience, denote  $K_1={K\sqrt{1-\alpha_2}\over 2}$.  Then
$$
\begin{array}{ll}
{e^{\alpha_1}\over (\sqrt{\pi})^n}\int_{|y|\geq K|x|}|y|^ie^{-{(1-\alpha_2)|y|^2\over 2}}\,dy&={e^{\alpha_1}\over (\sqrt{\pi})^n}\left({2\over \sqrt{1-\alpha_2}}\right)^{n+i}\int_{|z|\geq {K_1|x|}}|z|^ie^{-2|z|^2}\,dz\\[5mm]
&\leq {e^{\alpha_1}\over (\sqrt{\pi})^n}\left({2\over \sqrt{1-\alpha_2}}\right)^{n+3}\int_{|z|\geq {K_1|x|}}e^{-|z|^2}\,dz.
\end{array}
$$
The last inequality is due to $e^{{r^2}}\geq r^m$ if $r\geq J_m$ and $K_1|x|\geq J_m$.

 Note 
$$
\left(\int_{-r}^{r}e^{-t^2}\,dt\right)^2\geq \int_{\{{w\in \Rset^2|\ |y|\leq r}\}}e^{-|w|^2}\,dw=\pi\left(1-{e^{-r^2}}\right).
$$
Combining with $(1-t)^n\geq 1-nt$ for $t\in [0,1]$,  we deduce
$$
 \int_{Q_{r}}e^{-{|y|^2}}\,dy=\left(\int_{-r}^{r}e^{-t^2}\,dt\right)^n\geq (\sqrt{\pi})^{n}\left(1-{e^{-r^2}}\right)^{n\over 2}\geq (\sqrt{\pi})^n\left(1-{ne^{-r^2}\over 2}\right)
$$
Here $Q_r=\{y=(y_1,y_2,...,y_n)|\  |y_i|\leq r\ \text{for $i=1,2,...,n$}\}$. Accordingly, for $r\geq 1$, 
$$
\begin{array}{ll}
{1\over (\sqrt{\pi})^n}\int_{|y|\geq r}e^{-{|y|^2}}\,dy&\leq {1\over (\sqrt{\pi})^n}\int_{\Rset^n\backslash Q_r}e^{-{|y|^2}}\,dy+ {1\over (\sqrt{\pi})^n}\int_{Q_r\backslash \{|y|\leq r\}}e^{-{|y|^2}}\,dy\\[5mm]
&\leq {ne^{-r^2}\over 2}+{1\over (\sqrt{\pi})^n}e^{-r^2}(2r)^n< 2^ne^{-r^2}r^n
\end{array}
$$
This implies that
$$
{1\over (\sqrt{\pi})^n}\int_{|z|< {K_1|x|}}e^{-|z|^2}\leq 2^{n}e^{-K_1^2|x|^2}\left(K_1|x|\right)^n.
$$

Since $K_1\geq \max\{3\sqrt{\beta_1},  {2\sqrt{n\log n}\over |x|}\}$, 
$$
{e^{-2\beta_1|x|^2}}K_1^n |x|^ne^{-{K_1^2|x|^2}}\leq {e^{-2\beta_1|x|^2}}e^{-{K_1^2|x|^2\over 2}}\leq e^{-{K_1^2|x|^2\over 4}}
$$
Combining with 
$$
{K_1^2|x|^2\over 4}\geq {\tilde K^2|x|^2\over 4}\geq  (n+3)\log \left({1+4\beta_1\over \sqrt{1-\alpha_2}}\right)+16n\log n+\alpha_1+1+{{{\beta_2}^2\over \beta_1}},
$$
we obtain
$$
{e^{-2\beta_1|x|^2}}I_i\leq 2^ne^{\alpha_1}\left({2\over \sqrt{1-\alpha_2}}\right)^{n+3}e^{-{K_1^2|x|^2\over 4}}\leq   e^{-{{\beta_2}^2\over \beta_1}}(1+4\beta_1)^{-{n\over 2}}.
$$

By \eqref{eq4}, 
$$
\mathrm{I_i}\leq \bar p(t,x).
$$
  Hence   \eqref{all-in-one} holds.  As an immediate conclusion,  we have that
$$
|\nabla\bar q|={|\nabla \bar p|\over \bar p}\leq \beta_1|x|+\beta_2+2\beta_1\sqrt{2}K|x|< 3\beta_1K|x|+\beta_2.
$$

\medskip

Secondly,  to prove  \eqref{2ndorderestimate} and \eqref{testimate},  we first assume that $g(0)=0$ and $\nabla g(0)=0$, which will be removed at the end.  Then
$$
|\nabla g(x)|\leq L|x|.
$$

Next we first  verify  \eqref{2ndorderestimate}.  Note that
\be\label{Hessianexpression}
D^2\bar q=A-B+D\bar q\otimes D\bar q
\ee
Here
$$
A={1\over \bar p}{1\over {(\pi )}^{n\over 2}}\int_{\Rset^n}e^{-{|y|^2}}hD^2g(x-\sqrt{2t}y)\,dy
$$
and
$$
B={1\over \bar p}{1\over {(\pi )}^{n\over 2}}\int_{\Rset^n}e^{-{|y|^2}}hDg\otimes Dg(x-\sqrt{2t}y)\,dy.
$$
Here $(u\otimes u)_{ij}=u_iu_j$ is the outer product.  By Cauchy inequality,  $B\geq D\bar q\otimes D\bar q$. Hence
\be\label{H-2}
D^2\bar q\leq \mathrm{A}\leq LI_n. 
\ee
For the other direction, 
$$
D^2\bar q\geq {\mathrm{A}}-{\mathrm{B}}\geq -LI_n-\mathrm{B}. 
$$
So we just need to estimate the term B.  Note that
$$
\begin{array}{ll}
\left|\left|{{1\over {(\pi )}^{n\over 2}}\int_{\Rset^n}e^{-{|y|^2}}hDg\otimes Dg(x-\sqrt{2t}y)\,dy}\right|\right|_2&\leq {L^2\over {(\pi )}^{n\over 2}}\int_{\Rset^n}(|x|+\sqrt{2}|y|)^2e^{-{|y|^2}}h\,dy\\[5mm]
&\leq{2L^2\over {(\pi )}^{n\over 2}}\int_{\Rset^n}(|x|^2+{2}|y|^2)e^{-{|y|^2}}h\,dy\\[5mm]
&\leq 2L^2\left(|x|^2\bar p(t,x)+{2\over {\pi}^{n\over 2}}\int_{\Rset^n}|y|^2e^{-{|y|^2}}h\,dy\right).
\end{array}
$$

Thanks to \eqref{all-in-one} for $\beta_1=L$,  $\beta_2=0$ and $K=K_0$,   we have that
$$
||\mathrm{B}||_2\leq 2L^2(|x|^2+4K_0^2|x|^2)< 10L^2K_0^2|x|^2.
$$
Hence
$$
||D^2\bar q||_2\leq L+||B||_2<L+10L^2K_0^2|x|^2
$$

\medskip

Then let us verify \eqref{testimate}.  Note that
$$
\nabla \bar q_t(t,x)=C+D-\bar q_t\nabla \bar q.
$$
with
$$
\begin{array}{ll}
|C|&=\left|{1\over \sqrt{2t}}{1\over \bar p}{1\over {(\pi )}^{n\over 2}}\int_{\Rset^n}e^{-{|y|^2}}hD^2g(x-\sqrt{2t}y)\cdot y\,dy\right|\leq {L\over \sqrt{2t}}{1\over \bar p}{1\over {(\pi )}^{n\over 2}}\int_{\Rset^n}|y|e^{-{|y|^2}}h\,dy\\[5mm]
&\leq{2L\over \sqrt{2t}}K|x|<{2L\over \sqrt{t}}K|x|.
\end{array}
$$
Also, 
$$
\begin{array}{ll}
|D|=&\left|{1\over \sqrt{2t}}{1\over \bar p}{1\over {(\pi )}^{n\over 2}}\int_{\Rset^n}e^{-{|y|^2}}h\nabla g\otimes \nabla g(x-\sqrt{2t}y)\cdot y\,dy\right|\\[5mm]
&\leq{1\over \sqrt{2t}}{2L^2\over {(\pi )}^{n\over 2}}\int_{\Rset^n}(|x|^2|y|+{2}|y|^3)e^{-{|y|^2}}h\,dy\\[5mm]
&<{1\over \sqrt{t}}{2L^2|x|^2\over {(\pi )}^{n\over 2}}\int_{\Rset^n}|y|e^{-{|y|^2}}h\,dy+{1\over \sqrt{t}}{4L^2\over {(\pi )}^{n\over 2}}\int_{\Rset^n}|y|^3e^{-{|y|^2}}h\,dy\\[5mm]
&\leq{2L^2\over \sqrt{t}} (2K|x|^3+ 4K^3|x|^3)\\[5mm]
&<{12L^2\over \sqrt{t}}K^3|x|^3.
\end{array}
$$

Finally,
$$
\begin{array}{ll}
|\bar q_t|&=\left|{1\over \sqrt{2t}}{1\over \bar p}{1\over {(\pi )}^{n\over 2}}\int_{\Rset^n}e^{-{|y|^2}}h\nabla g(x-\sqrt{2t}y)\cdot y\,dy\right|\\[5mm]
&\leq {L\over \sqrt{2t}}{1\over \bar p}{1\over {(\pi )}^{n\over 2}}\int_{\Rset^n}(|x||y|+\sqrt{2}|y|^2)e^{-{|y|^2}}h\,dy\\[5mm]
&\leq {L\over \sqrt{2t}}(2K|x|^2+ 2\sqrt{2}K^2|x|^2)\leq  {6L\over \sqrt{t}}K^2|x|^2.
\end{array}
$$
Also,
$$
|\nabla \bar q|\leq 2L(1+\sqrt{2})K|x|<6LK|x|.
$$

Hence
$$
|\bar q_t\nabla q|\leq {36L^2 \over \sqrt{t}}K^3|x|^3.
$$
So
$$
\begin{array}{ll}
|\nabla \bar q_t(t,x)|&\leq {2L\over \sqrt{t}}K|x|+{1\over \sqrt{t}}12L^2K^3|x|^3+{36L^2 \over \sqrt{t}}K^3|x|^3\\[5mm]
& \leq   {50L^2\over \sqrt{t}}K^3|x|^3.
\end{array}
$$

At last, for a general $g$,  we consider
$$
g_0(x)=g(x)-g(0)-\nabla g(0)\cdot x.
$$
and let $\bar q_0(t,x)$ be the corresponding score function.  Then we have 
$$
g_0(0)=0, \quad  \nabla g_0(0)=0\quad \mathrm{and} \quad \bar q_0(t,x)=\bar q(t,x+\nabla g(0)).
$$
Hence \eqref{2ndorderestimate} and  \eqref{testimate} hold for general cases. 
\qed

\begin{rmk}\label{moreont} In the proof of \eqref{testimate}, we may also bound $\bar q_t$ using the equation
$$
\bar q_t={\frac{1}{2}}(\Delta \bar q-|\nabla \bar q|^2)
$$
together with bounds for $\Delta \bar q$ and $|\nabla \bar q|^2$. Note $\nabla g(x-\sqrt{t}y)=\nabla g(x)+r_t$ for $|r_t|\leq  L\sqrt{t}|y|$.   Then the term $D$ in the proof
\begin{align*}
\begin{array}{ll}
D&={1\over \sqrt{2t}}{1\over \bar p}{1\over {(\pi )}^{n\over 2}}\int_{\Rset^n}e^{-{|y|^2}}h\nabla g\otimes \nabla g(x-\sqrt{2t}y)\cdot y\,dy\\
&=\bar q_t\nabla g(x)+O(L^2K^3|x|^3).
\end{array}
\end{align*}
This will lead to a bound of  $\nabla \bar q_t$ by replacing $L^2\over \sqrt{t}$ in  \eqref{testimate} by $O(nL^3)$.  The details are left to interested readers as an exercise. 
\end{rmk}

Below are other simple situations where we can obtain a global uniform bound of the Hessian, which follows immediately from \eqref{gradientexpression} and \eqref{Hessianexpression}. 
\begin{theo}\label{othercondition} Let $L_1$ and $L_2$ be two positive constants such that 
$$
|\nabla g|\leq L_1  \quad \mathrm{and} \quad ||D^2g||_{2}\leq L_2.
$$
Then 
$$
|\nabla \bar q(t,x)|\leq L_1 \quad \mathrm{and} \quad -(L_2+L_1^2)I_n\leq  D^2q(t,x)\leq L_2I_n
$$
\end{theo}
\subsection{Proof of Theorem \ref{linearcontrol2}}\label{proof:linearcontrol2} We first prove (1).  Note that $D\bar q=-{D\bar p\over \bar p}$ and
$$
 \begin{array}{ll}
\bar p(t,x)&={1\over {(2\pi t)}^{n\over 2}}\int_{\Rset^n}e^{-{|x-y|^2\over 2t}}d\pi_0(y)\\
&={1\over {(2\pi t)}^{n\over 2}}\int_{D_0}e^{-{|x-y|^2\over 2t}}d\pi_0(y).
\end{array}
$$
Then
$$
\nabla\bar p(t,x)=-{1\over {(2\pi t)}^{n\over 2}}\int_{D_0}{(x-y)\over t}e^{-{|x-y|^2\over 2t}}d\pi_0(y).
$$
Since $D_0\subset {\overline B_M(0)}$
$$
|\nabla\bar p(t,x)|\leq {|x|+M\over t}\bar p(t,x).
$$

Next  we prove \eqref{l-bound}.  Note that
$$
-D^2\bar q(t,x)={D^2\bar  p\over \bar p}-{\nabla \bar p\otimes\nabla \bar p\over\bar  p^2}=-{\delta_{ij}\over t}+{1\over t^2}{A-B\over \hat p^2(t,x)}.
$$
Here $\hat p(t,x)=\int_{D_0}e^{-{|x-y|^2\over 2t}}d\pi_0(y)$, 
$$
A_{ij}=\hat p(t,x)\int_{D_0}(x_i-y_i)(x_j-y_j)e^{-|x-y|^2\over 2t}d\pi_0(y)
$$
and
$$
\begin{array}{ll}
B_{ij}&=\int_{D_0}(x_i-y_i)e^{-|x-y|^2\over 2t}d\pi_0(y)\cdot \int_{D_0}(x_j-y_j)e^{-|x-y|^2\over 2t}d\pi_0(y)\\
&=\hat p^2(t,x)x_ix_j- \hat p(t,x)\int_{D_0}(x_iy_j+x_jy_i)e^{-|x-y|^2\over 2t}d\pi_0(y)+\\
&\quad +\int_{D_0}y_ie^{-|x-y|^2\over 2t}d\pi_0(y)\cdot \int_{D_0}y_je^{-|x-y|^2\over 2t}d\pi_0(y)
\end{array}
$$
Hence $(A-B)_{ij}$
\be\label{abbound}
\hat p(t,x)\int_{D_0}y_iy_je^{-|x-y|^2\over 2t}d\pi_0(y)-\int_{D_0}y_ie^{-|x-y|^2\over 2t}d\pi_0(y)\cdot \int_{D_0}y_je^{-|x-y|^2\over 2t}d\pi_0(y)
\ee
So it is easy to see that 
$$
||A-B||_2\leq  M^2\hat p^2(t,x).
$$
Thus \eqref{l-bound} holds. \qed

\subsection{Proof of Theorem \ref{theo:optimalbound}}\label{prof:theo:optimalbound}  Since $\pi_0$ is comparable to the uniform distribution, there exists a constant $C$ such that for any measurable subset $U\subset D_0$
\be\label{measure-compare}
{1\over C}\mathcal{H}_d(U)\leq \pi_0(S)\leq C \mathcal{H}_d(U)
\ee
 Here $\mathcal{H}_d(\cdot)$ is the $d$-dimensional Hausdorff measure. Hereafter, we write 
 
 \begin{enumerate}[label=(\roman*)]
 
\item  $\partial D_0$:  the $d-1$ dimensional boundary of $D_0$.  Moreover, for $y\in D_0$;

\item  $T_y(D_0)\subset \Rset^n$:  the $d$-dimensional tangent space of $D_0$ at $y$;

\item $N_y(D_0)\subset \Rset^n$:   the  $n-d$ dimensional orthogonal complement of $T_y(D_0)$;

\item  $T'_y(D_0)\subset \Rset^n$:   the $d-1$-dimensional tangent space of $\partial D_0$ at $y\in \partial D_0$.  Note that  $T'_y(D_0)$ is a subspace of $T_y(D_0)$;

\item $N'_y(D_0)\subset \Rset^n$:   the  $n+1-d$ dimensional orthogonal complement of $T'_y(D_0)$.  $N_y(D_0)$ is a subspace of $N'_y(D_0)$ 
  
   \end{enumerate}

    Let 
$$
S=\{x\in \Rset^n|\  \text{there exists a unique $y_x\in D_0$ such that $|x-y_x|=d(x,D_0)$}\}.
$$
Then $\Rset^n\backslash S$ has zero measure since $d(x,D_0)$ is differentiable almost everywhere.

Write
$$
S_1=\{x\in S|\  y_x\in  D_0\backslash \partial D_0\} \quad  \mathrm{and} \quad S_2=\{x\in S|\  y_x\in  \partial D_0\},
$$
$$
W_1=\left\{x\in S_1|\  \liminf_{y\in D_0\to y_x} {|x-y|^2-|x-y_x|^2\over |y-y_x|^2}>0\right\},
$$
$$
W_2=\left\{x\in S_2|\ x-y_x\in N_{y_x}(D_0) \quad \mathrm{and} \quad  \liminf_{y\in D_0\to y_x} {|x-y|^2-|x-y_x|^2\over |y-y_x|^2}>0\right\},
$$
and
$$
W_3=\left\{x\in S_2|\ x-y_x\notin N_{y_x}(D_0) \quad \mathrm{and} \quad  \liminf_{y\in \partial D_0\to y_x} {|x-y|^2-|x-y_x|^2\over |y-y_x|^2}>0\right\}.
$$
Note that 
$$
\begin{array}{ll}
\text{ if $y\in D_0\backslash \partial D_0$}, \quad \text{ then  $x-y_x\in N_{y_x}(D)$}, \\
\text{ if $y\in \partial D_0$}, \quad \text{then  $x-y_x\in N'_{y_x}(D)$}.
\end{array}
$$

{\bf Step 1:}  We show that $S\backslash \left( \cup_{i=1}^{3}W_i\right)=(S_1\backslash W_1)\cup (S_2\backslash W_2)\cup (S_2\backslash W_3)$ has zero measure.    We will prove this for $S_1\backslash W_1$, The proofs for the other two are similar.   Apparently,  if $x\in S$, then for all $t\in (0,1)$, $y_{x_t}=y_x$ and $x_t\in W$ for $x_t=y_x+t(x-y_x)$.   Also,
for $y\in D_0\backslash \partial D_0$, let us write
$$
\Gamma_y=\{x\in S_1\backslash W_1|\  y_x=y\}.
$$
By compactness argument, it is easy to show that  for  given $x\in S$ and $r>0$,  there exists a $r_x>0$, such that $y_{\tilde x}\in B_r(y_x)$ for any $\tilde x\in B_{r_x}(x)\cap S$ .  Hence to prove that $S_1\backslash W$ has zero measure, it suffices to show that  for any $y_0\in D_0$,  if $\Gamma_{y_0}$ is not empty, then there exists $r_0>0$ such that 
\be\label{L-graph}
\cup_{y\in B_{r_0}(y_0)\cap D_0}\Gamma_{y}\subset  \{y+t(y,v)|\ y\in B_{r_0}(y_0),\ v\in N_y(D_0) \ \mathrm{and}\ |v|=1\}
\ee
for a locally Lipschitz continuous function 
$$
t(y,v): B_{r_0}(y_0)\times  N_y(D_0)\to (0,\infty).
$$

  By suitable translation and rotation, we may assume  $y_0=0$ and in a neighbourhood $V$ of $0$,
\be\label{coordinate1}
D_0\cap V=V\cap \{(y',F(y'))|\  y'\in \Rset^{d}\},
\ee
where  $F=(F^{(d+1)},F^{(d+2)},..., F^{(n)}):\Rset^d\to \Rset^{n-d}$ is a smooth map satisfying $\nabla F(0)=0$ and $F(0)=0$.  Choose $x\in \Gamma_{y_0}$.   Let $v={x\over |x|}\in N_y(D_0)$ and $t>0$, let
$$
H_{tv}(y')=|tv-(y',F(y'))|^2.
$$
Since $x\in W_1$,  $D^2H_{tv}(0)$ cannot be a positive definite matrix for $t=|x|$.  Meanwhile,  for $0<t<|x|$,  $y_{tx}=y_0$ and $tx\in W_1$, which implies that  $D^2H_{tv}(0)$ is positive definite for $t\in [0, |x|)$.  Hence $t=|x|$ is the first moment such that $D_{tv}^2H(0)$ has a zero eigenvalue, 
which is equivalent to saying that 
$$
\text{the largest eigenvalue of the $d\times d$ matrix $\sum_{k=d+1}^{n}tv_kF^{(k)}_{y'_iy'_j}(0)$ is 1.}
$$
Therefore, for $y\in B_r(y_0)$ and $v\in T_y(D_0)$ with $|v|=1$, if  the largest eigenvalue $\lambda (y,v)$ of  the matrix $\sum_{k=d+1}^{n}v_kF^{(k)}_{y'_iy'_j}(y)$ is  positive, we set
$$
t(y,v)={1\over \lambda (y,v)}.
$$
Then \eqref{L-graph} holds.

{\bf Step 2:}  We will verify that  if $x\in \cup_{i=1}^{3}W_i$, then
$$
|D^2\bar q(t,x)|\leq {C_x\over t} \quad \text{for all $t\in (0,1]$}. 
$$

Case 1:  Assume that $x\in W_1$.  Without loss of generality,  we may assume $y_x=0$ and use the representation as \eqref{coordinate1}.  Choose $r>0$ such that

(i)$V_r= \{(y',F(y'))|\  |y'|<r\}\subset D_0$;

(ii) Then there exists $\alpha_x, \beta_x>0$ such that for $y\in V_r$, 
\be\label{theothersidesquarebound}
\alpha_x|y-y_x|^2\geq |x-y|^2-|x-y_x|^2\geq \beta_x |y-y_x|^2 \quad \text{for $y\in D_0$}.
\ee
For $k\geq 1$ and $k\sqrt{t}\leq r$,  write 
$$
V_{t,k}= \left\{(y',F(y'))|\  |y'|<k\sqrt{t}\right\}
$$
Thanks to  the left upper bound in  \eqref{theothersidesquarebound},
$$
\int_{D_0}e^{-|x-y|^2\over 2t}\,d\pi_0\geq \int_{V_{t,1}}e^{-|x-y|^2\over 2t}\,d\pi_0\geq O\left(t^{d\over 2}e^{-{|x-y_x|^2\over 2t}}\right).
$$
Recall that $y_x=0$.   To see the dependence on $y_x$, we keep $y_x$ in the computations below instead of replacing it by $0$.   Note
$$
\int_{D_0}|y-y_x|^2e^{-|x-y|^2\over 2t}\,d\pi_0\leq \int_{V_r}|y-y_x|^2e^{-|x-y|^2\over 2t}\,d\pi_0+Ce^{-{{|x-y_x|^2+\delta_r}\over 2t}}
$$
for some $\delta_r>0$.  

Also,
$$
\begin{array}{ll}
&\int_{V_r}|y-y_x|^2e^{-|x-y|^2\over 2t}\,d\pi_0=\sum_{k=0}^{\infty}\int_{\{y\in V_{t,k+1}\backslash V_{t,k}\}}|y-y_x|^2e^{-|x-y|^2\over 2t}\,d\pi_0\\
&\leq Ct\cdot  t^{d\over 2}e^{-{|x-y_x|^2\over 2t}}\sum_{k=1}^{\infty} (k+1)^2e^{-k}=O\left(t^{d\over 2}e^{-{|x-y_x|^2\over 2t}}\right)t.
\end{array}
$$
Hence
\be\label{distancetocenter}
{\int_{D_0}|y-y_x|^2e^{-|x-y|^2\over 2t}d\pi_0(y)\over \hat p}\leq Ct
\ee
Recall that
$$
\hat p(t,x)={\int_{D_0}e^{-|x-y|^2\over 2t}d\pi_0(y)}=(\sqrt{2\pi t})^{n}\bar p(t,x).
$$
Then for $1\leq i,j\leq n$ and $y_x=(a_1,a_2,...,a_n)$, 
$$
\begin{array}{ll}
&\left|{\int_{D_0}y_iy_je^{-|x-y|^2\over 2t}d\pi_0(y)\over \hat p(x,t)}-{\int_{D_0}y_ie^{-|x-y|^2\over 2t}d\pi_0(y)\over \hat p(x,t)}\cdot {\int_{D_0}y_je^{-|x-y|^2\over 2t}d\pi_0(y)\over \hat p(x,t)}\right|= \\
&\left|{{\int_{D_0}(y_i-a_i)(y_j-a_j)e^{-|x-y|^2\over 2t}d\pi_0(y)\over \hat p(x,t)}}-{\int_{D_0}(y_i-a_i)e^{-|x-y|^2\over 2t}d\pi_0(y)\over \hat p(x,t)}\cdot {\int_{D_0}(y_j-a_j)e^{-|x-y|^2\over 2t}d\pi_0(y)\over \hat p(x,t)}\right|\\
&\leq Ct. 
\end{array}
$$
The last equality follows from  \eqref{distancetocenter} and the Cauchy inequality.  Therefore, \eqref{abbound} leads to 
$$
|D^2\bar q(x,t)|\leq {C_x\over t}.
$$

Case 2: $x\in W_2$.  The proof is similar to Case 1.

Case 3:  $x\in W_3$.   By suitable translation and rotation, we may assume  $y_x=0$ and in a neighborhood  of $0\in \Rset^d$,
$$
D_0\cap V= \tilde V_r=\{(y',F(y'))|\ y'=(y'_1,...,y'_d)\in  \Omega_{f,r}\},
$$
where  $F:\Rset^d\to \Rset^{n-d}$ is smooth map satisfying that $\nabla F(0)=0$. Also,  
$$
\Omega_{f,r}=\{z=(z', z_d)|\ z'=(z_1,z_2,..,z_{d-1})\in \Rset^{d-1},\ |z'|<r; \;
z_d\geq f(z_1,z_2,..,z_{d-1})\}.
$$
for a smooth function  $f:\Rset^{d-1}\to \Rset$ subject to $\nabla f(0)=0$.  Then
$$
T_0(D_0)=\{(v,0,...,0)\in \Rset^n\ |\ v\in \Rset^d\}
$$
and the $d-1$ dimensional tangent plane to $\partial D_0$ at $y_x=0$ is 
$$
\partial T_0(D_0)=\{(v',0,0,...,0)\in \Rset^n\ |\ v'\in \Rset^{d-1}\}.
$$
Thus
$$
x=x-y_x=( \underbrace{0,...0}_{d-1}, \theta_x, z_x)
$$
for some $\theta_x>0$ and $z_x\in  \Rset^{n-d}$.   To see the dependence on $y_x$, as in Case 1, we keep $y_x$ in the computations below instead of replacing it by $0$.

Then for $y''\in \Rset^{d-1}$ and $y=(y'',y_{d},  F(y'',y_d))\in D_0$, 
\be\label{generalbound}
|x-y|^2-|x-y_x|^2= 2\theta_x(y_d-f(y'))+O(|y-y_x|^2).
\ee

Write
$$
H(y)=|x-y|^2-|x-y_x|^2=H(y'',y_d,F(y'',y_d)).
$$

Since $x\in W_3$, 
$$
H(y'',f(y''),F(y'',f(y'')))\geq  \delta_x|y''|^2. 
$$

Therefore,   there exists $r>0$ and $M>0$ such that
\be\label{positivebound}
|x-y|^2-|x-y_x|^2\geq {\theta_x\over M} (y_d-f(y'))+\delta_x|y-y_x|^2 \quad \text{for all $y\in \tilde V_r$}.
\ee

Write
$$
R_{t,k}=\{(y',y_d, F(y', y_d))\in \Omega|\   |y'|\leq k\sqrt{t} \quad \mathrm{and} \quad  0\leq y_d-f(y')\leq kt\}.
$$
Thanks to \eqref{generalbound}, 
$$
\hat p(t,x)=\int_{D_0}e^{-|x-y|^2\over 2t}\,d\pi_0\geq \int_{D_0\cap R_{t,1}}e^{-|x-y|^2\over 2t}\,d\pi_0\geq O\left(t^{d+1\over 2}e^{-{|x-y_x|^2\over 2t}}\right).
$$

Note that 
$$
\int_{D_0}|y-y_x|^2e^{-|x-y|^2\over 2t}\,d\pi_0\leq \int_{\tilde V_r}|y-y_x|^2e^{-|x-y|^2\over 2t}\,d\pi_0+Ce^{-{|x-y_x|^2+\delta_r\over 2t}}.
$$
Also, 
$$
\begin{array}{ll}
&\int_{\tilde V_r}|y-y_x|^2e^{-|x-y|^2\over 2t}\,d\pi_0=\sum_{k=0}^{\infty}\int_{D_0\cap( R_{t,k+1}\backslash R_{t,k})}|y-y_x|^2e^{-|x-y|^2\over 2t}\,d\pi_0\\
&\leq C t\cdot t^{d+1\over 2}e^{-{|x-y_x|^2\over 2t}}\sum_{k=1}^{\infty} (k+1)^2e^{-k}=tO\left(t^{d+1\over 2}e^{-{|x-y_x|^2\over 2t}}\right).
\end{array}
$$
Hence 
$$
{\int_{D_0}|y-\bar y_x|^2e^{-|x-y|^2\over 2t}d\pi_0(y)\over \hat p}\leq Ct
$$
Then by the same argument in the end of Case 1, we deduce that
$$
||D^2\bar q(x,t)||_2\leq {C_x\over t}.
$$

Finally, if $D_0$ is convex, then  it is clear that $S=\Rset^n$ and $W_1=S_1$ and $W_2\cup W_3=S_2$.  Hence the $O({1\over t})$ bound holds for all $x\in \Rset^n$. 
 \qed

\subsection{Proof of Example \ref{eg:Hessian-growth-t2}}\label{proof:eg:Hessian-growth-t2}
Proof: For given $x\in \Rset^2$,  denote by $\bar y_t$  the weighted center of mass:
$$
\bar y_t={\int_{D_0}ye^{-|x-y|^2\over 2t}d\pi_0(y)\over \hat p}
$$
Note that as $t\to 0$, the measure $e^{-|x-y|^2\over 2t}d\pi_0(y)\over \hat p$ will concentrate on $\{y\in D_0|\ |y-x|=d(x, D_0)\}$.  Thus, 
$$
\lim_{t\to 0}d(\bar y_t, \Gamma_x)=0,
$$
where $\Gamma_x$ is the convex hull of $\{y\in D_0|\ |y-x|=d(x, D_0)\}$.   According to the computation in the proof of \eqref{l-bound}, we have that
\begin{align}
   -\Delta \bar q=&-{n\over t}+{\hat p\int_{D_0}|y|^2e^{-|x-y|^2\over 2t}d\pi_0(y)-\left|\int_{D_0}ye^{-|x-y|^2\over 2t}d\pi_0(y)\right|^2\over t^2\hat p^2}\\
   =&-\Delta \bar q=-{n\over t}+{\int_{D_0}|y-\bar y_t|^2e^{-|x-y|^2\over 2t}d\pi_0(y)\over t^2\hat p},
\end{align}
where the second term is like a variance. 
If $x=(\theta,0)$ for some $\theta>0$, there are two points $y_1$ and $y_2$ such that
$$
|x-y_1|=|x-y_2|=d(x,D_0). 
$$
Due to the symmetry,  we must have that 
$$
{e^{-|x-y|^2\over 2t}d\pi_0(y)\over \hat p}\to {\frac{1}{2}}\delta_{y_1}+{\frac{1}{2}}\delta_{y_2} \quad \mathrm{and} \quad \lim_{t\to +\infty}y_t={y_1+y_2\over 2}.
$$
Accordingly, 
$$
\lim_{t\to 0}{\int_{D_0}|y-\bar y_t|^2e^{-|x-y|^2\over 2t}d\pi_0(y)\over \hat p}={|y_1-y_2|^2\over 4},
$$
leading to 
$$
-\Delta \bar q(x,t)\geq {C_x\over t^2} \quad \text{for $t\in (0,1]$.} \quad \quad \quad \qed
$$

 \subsection{Proof of Theorem \ref{well-posed}}\label{proof:well-posed}
\begin{lem}\label{existence-ode} Given $T>0$,   suppose that $F=F(t,x)\in C([0,T]\times \Rset^n, \Rset^n)$ satisfies that $F$ is locally Lipschitz continuous in $x$ variable, i.e., for any $M>0$, there exists a constant $L_M$ such that
$$
|F(t,x)-F(t,y)|\leq L_M|x-y| \quad \text{for $x,y\in B_M(0)$ and $t\in [0,T]$}
$$
and
\be
|F(t,x)|\leq C(|x|+1).  \quad \text{for $(t,x)\in  [0,T]\times \Rset^n$}.
\ee
for a positive constant C.  Then for any $x_0\in \Rset^n$, the following equation has a unique solution
$$
\begin{cases}
\dot X(t)=F(t,X(t)) \quad \text{$t\in [0,T]$}\\
X(0)=x_0.
\end{cases}
$$
\end{lem}

Proof: The uniqueness follows from standard ODE theory.  We just need to establish the global existence.  Let $w(t)=|X(t)|^2$. Then
$$
\dot w(t)\leq C_1w(t)+C_2
$$
for two positive constants $C_1$ and $C_2$ depending only on $C$.  Hence  for all $t\geq 0$,
$$
e^{-C_1t}w(t)\leq |x_0|^2+{C_2\over C_1}\left(1-e^{-C_1t}\right).
$$
Hence the solution can be extended to $T$.  \qed

\paragraph{Proof of Theorem \ref{well-posed}}
It suffices to notice that  for each fixed sample $\omega$,  $Y(t)=Y(t,\omega)=X_t(\omega)-W_t(\omega)$ just satisfies the regular ODE for any fixed sample
$$
\begin{cases}
dY(t)=F(Y+W(t),t)dt \quad \text{$t\in [0,\infty)$}\\
Y(0)=x_0.
\end{cases}
$$
Hence the Corollary follows from Theorem \ref{existence-ode} and the well-known fact that $W_t(\omega)\in C([0,T], \Rset^n)$  for a.e. $\omega$. 
\qed
\subsection{ Proof of Theorem \ref{thm:conv-1}}\label{proof:conv-1}
The key ingredient is the following estimates on truncation error.
\begin{theo}\label{thm:accumuated-error} Under the same assumption as Corollary \ref{cor:n-control}, we have that for fixed $T\leq 1$ and $t_k={kT\over N}$, 
\be\label{accumuated-error}
\sum_{i=1}^{N}\int_{t_{k-1}}^{t_{k}}\mathbb{E}||\nabla \bar q(t_k,x(t_k))-\nabla \bar q(t,x(t))||^2\,dt\leq {CL^6Tn(n\log n)^2\over N}
\ee
Here $C$ is a constant independent of $n$ and $L$.

\end{theo}

\textbf{Proof:} It suffices to show that for $s>t\in [0,T]$, 
$$
\mathbb{E}||\nabla \bar q(s, x(s))-\nabla \bar q(t,x(t))||^2\leq CL^6(s-t)n^3\log n
$$
According to Lemma C.6 in \cite{chen2023improved},  it suffices to show that
\be\label{n-square-bound}
\mathbb{E}||\nabla \bar q(t, x(t)+z)-\nabla \bar q(t,x(t))||^2\leq Cn^2(s-t).
\ee
Here $z \thicksim \mathcal{N}(0, C(s-t))$.

 Owing to Theorem \ref{linearcontrol1} in our paper,
$$
||D^2\bar q(t,x)||_2\leq C(|x|^2+n\log n).
$$
See \eqref{matrixnorm}  for the definition of the spectral norm $||\cdot||_2$ of $n\times n$ matrix. Then
$$
||\nabla \bar q(t,x(t)+z)-\nabla \bar q(t,x(t))||^2\leq C(1+|x(t)|^4+|z|^4+(n\log n)^2)|z|^2.
$$
Note that $\mathbb{E}(z^2)\leq Cn(s-t)$ and $\mathbb{E}(z^4)\leq Cn^2(s-t)^2$. Moreover,  by Cauchy inequality
$$
\mathbb{E}(|x(t)|^4z^2)\leq \sqrt{(\mathbb{E}(x^8(t))\mathbb{E}(z^4)}\leq Cn(n\log n)^2(s-t).
$$
The last inequality is due to $\mathbb{E}_{p(t,x)}(x(t)^8)\leq C(n\log n)^4$ from Remark \ref{n-control}.   Hence \eqref{n-square-bound}holds. \qed

\begin{rmk} In the proof of Theorem \ref{thm:accumuated-error},  instead of using Lemma C.6 in \cite{chen2023improved},  we may also use \eqref{testimate} from Theorem \ref{linearcontrol1}  to bound the difference between two times and so:
$$
\mathbb{E}||\nabla \bar q(s, x(t))-\nabla \bar q(t,x(t))||^2\leq C(s-t)^2n^4(\log n)^3.
$$
This leads to an extra term ${Cn^4(\log n)^3\over N^2}$ on the right hand side of \eqref{accumuated-error}.  The proof is similar. Note that when $N=O(n^2)$,  ${n^4(\log n)^3\over N^2}\preceq{n^3(\log n)^2\over N}$. 
\end{rmk}
\begin{rmk}
    We are aware of the difference of $\nabla \log p$ and $\nabla \bar q$ due to the translation \eqref{transform}, while our Lipschitz estimate is uniform in time, hence similar results of Theorem \ref{thm:accumuated-error} hold for $\nabla \log p$.
\end{rmk}
\subsection{Convergence bounds under compact support manifold assumption}\label{sec:conv-compact}
\begin{theo}
     Assume that $\mathrm{supp}(p_0)=D_0\subset \overline {B_M(0)}$ and the density is smoothly defined on $D_0$. Consider early stopping, so let $\delta>0$, and  let $\hat{q}_{T-\delta}$ be the distribution generated by the uniform discretization scheme of the exponential integrator  \eqref{semi-discrete}, with an approximated score satisfying the Assumption \ref{ass:score}.
     If $L_1> 0$, then 
    \begin{align*}
\mathrm{KL}(p_\delta\| \hat{q}_{T-\delta}) \lesssim (M_2+d)e^{-T}+T\epsilon_0^2+\frac{dT^2L^2_\delta}{N},
\end{align*}
where $L_\delta =1+\frac{1}{\delta}+\frac{M^2}{\delta^2}$.
\end{theo}
\textbf{Proof:}
The $L_\delta$ is computed from Theorem \ref{linearcontrol2}. Then Proposition \ref{est-global-lip} applies and yields the proof. \qed
 
\section{Broader Impact}\label{broaderimpact}
Diffusion model is one of the most influential generative models in the era of AI. Our theory provides theoretical guarantee of diffusion models with minimal assumption on data distribution. In addition, we provide a theoretical characterization of the singular behaviour related to the manifold hypothesis. This  provides insight for model parameterization in practical implementations.

\bibliographystyle{plain}
\bibliography{refs}

\end{document}